\pdfoutput=1
\documentclass[runningheads]{llncs}
\usepackage{graphicx}

\usepackage{tikz}
\usepackage{comment}
\usepackage{amsmath,amssymb} 
\usepackage{color}

\usepackage{multirow}
\usepackage{subfigure} 
\usepackage{booktabs}
\usepackage{algorithm}
\usepackage{algorithmic}
\newcommand{\expand}{\mbox{EXPAND}}

\begin{document}
\pagestyle{headings}
\mainmatter
\def\ECCVSubNumber{1825}  

\title{A Generic Graph-based Neural Architecture Encoding Scheme for Predictor-based NAS}


\titlerunning{GATES}
%

\author{Xuefei Ning\inst{1} \and
  Yin Zheng\inst{2} \and
Tianchen Zhao\inst{3} \and
Yu Wang\inst{1} \and
Huazhong Yang\inst{1}}
\authorrunning{X. Ning et al.}
%

\institute{Department of Electronic Engineering, Tsinghua University \and Weixin Group, Tencent \and Department of Electronic Engineering, Beihang University \\
\email{foxdoraame@gmail.com, yu-wang@tsinghua.edu.cn}}
\maketitle

\begin{abstract}
This work proposes a novel Graph-based neural ArchiTecture Encoding Scheme, a.k.a. GATES, to improve the predictor-based neural architecture search.
Specifically, different from existing graph-based schemes, GATES models the operations as the transformation of the propagating information, which mimics the actual data processing of neural architecture.
  GATES is a more reasonable modeling of the neural architectures, and can encode architectures from both the ``operation on node'' and ``operation on edge'' cell search spaces consistently. 
Experimental results on various search spaces confirm GATES's effectiveness in improving the performance predictor. Furthermore, equipped with the improved performance predictor, the sample efficiency of the predictor-based neural architecture search (NAS) flow is boosted. Codes are available at \url{https://github.com/walkerning/aw\_nas}.

\end{abstract}

\section{Introduction}
\label{sec:intro}

Recently, Neural Architecture Search (NAS) has received extensive attention due to its capability to discover neural network architectures in an automated manner. 
Substantial studies have shown that the automatically discovered architectures by NAS are able to achieve
highly competitive performance. 

Generally speaking, there are two key components in a NAS framework, the 
\textit{architecture searching module} and the \textit{architecture evaluation module}. Specifically, the architecture evaluation module provides the signals of the architecture performance, e.g., accuracy, latency, etc., which are then used by the architecture searching module to explore architectures in the search space.
In the seminal work of \cite{zoph2016neural}, the architecture evaluation is conducted by training every candidate architecture until convergence, 
and thousands of architectures need to be evaluated during the architecture search process. As a result, the computational burden of the whole NAS process is extremely large. There are two directions to address this issue, which focus on
improving the searching and evaluation module, respectively. 1) Evaluation: accelerating the evaluation of each individual architecture, and in the meanwhile, keep the evaluation meaningful in the sense of ranking correlation; 2) Searching: increasing the sample efficiency so that fewer architectures are needed to be evaluated for discovering a good architecture.

To improve the sample efficiency of the architecture searching module, a promising idea is to learn an approximated performance predictor, and then utilize the predictor to sample architectures that are more worth evaluating. We refer to these NAS methods~\cite{liu2018progressive,nao2018,wang2018alphax} as the predictor-based NAS methods, and their general flow will be introduced in Sec.~\ref{sec:pb-nas}. The generalization ability of the predictor is crucial to the sample efficiency of predictor-based NAS flows. Our work follows the line of research of predictor-based NAS, and focus on improving the performance predictor of neural architectures.

A performance predictor predicts the performance of architectures based on the encoding of them. Existing neural architecture encoding schemes include the sequence-based scheme and the graph-based scheme. The sequence-based schemes~\cite{nao2018,liu2018progressive,wang2018alphax} rely on specific serialization of the architecture. 
They model the topological information only implicitly, which deteriorates the representational power and interpretability of the predictor.
Existing graph-based schemes~\cite{guo2019nat,shi2019multi} usually apply graph convolutional networks (GCN)~\cite{kipf2016semi} to encode the neural architectures.
For the ``operation on node'' (OON) search spaces, in which the operations (e.g., \texttt{Conv3x3}) are on the nodes of the directed acyclic graph (DAG), GCN can be directly applied to encode architectures.
Nevertheless, since a neural architecture is a ``data processing'' graph, where the operations behave as the data processing functions (e.g., \texttt{Conv3x3}, \texttt{MaxPool}), existing methods' modeling of operations as the node attributes in OON search spaces is not suitable. Instead of modeling the operations as node attributes, a more natural solution is to treat them as the transforms of the node attributes (i.e., mimic the processing of the information).
On the other hand, for the ``operation on edge'' (OOE) search spaces,\footnote{Figure~\ref{fig:two_search_space} illustrates the OON and OOE search spaces.}
the handling of edge information in the existing graph-based scheme~\cite{guo2019nat} is even more unsatisfying regarding its poor generalizability and flawed handling of architecture isomorphism. 

In this work, we propose a general encoding scheme: \emph{Graph-based neural ArchiTecture Encoding Scheme (GATES)}, which is suitable for the representation learning of data processing graphs such as neural architectures.
Specifically, to encode a neural architecture, GATES models the information flow of the actual data processing of the architecture. First, GATES models the input information as the attributes of the input nodes. And the input information will be propagated along the architecture DAG. The data processing of the operations (e.g., \texttt{Conv3x3}, \texttt{MaxPool}) are modeled by GATES as different transforms of the information. Finally, the output information is used as the embedding of the cell architecture.
Since the encoding process of GATES mimics the actual computation flow of the architectures, GATES intrinsically maps isomorphic architectures to the same representation. Moreover, GATES can encode architectures from both the OON and OOE 
cell search spaces in a consistent way.
Due to the superior representational ability of GATES, the generalization ability of the architecture performance predictor using GATES is significantly better than other encoders.
Experimental results confirm that GATES is effective in improving the architecture performance predictors. Furthermore, by utilizing the improved performance predictor, the sample efficiency of the NAS process is improved.

\section{Related Work}
\label{sec:related_work}

\subsection{Architecture Evaluation Module}

One commonly used technique to accelerate architecture evaluation is parameter sharing~\cite{pham2018efficient}, where a super-net is constructed such that all architectures in the search space share a superset of weights and the training costs of architectures are amortized to an ``one-shot'' super-net training. Parameter sharing dramatically reduces the computational burden and is widely used by recent methods. 
However, recent studies~\cite{sciuto2019evaluating,luo2019understanding} find that the ranking of architecture candidates with parameter sharing does not reflect their true rankings well,
which dramatically affects the effectiveness of the NAS algorithm. Moreover, the parameter sharing technique is not generally applicable, 
since it is difficult to construct the super-net for some search spaces, for example, in NAS-Bench-101~\cite{ying2019bench}, 
one operation can have different output dimensions in different candidate architectures.
Due to these limitations, this work does not use the parameter sharing technique, and focus on improving the sample efficiency of the architecture searching module.

\subsection{Architecture Searching Module}
To improve the sample efficiency of the architecture search module, a variety of search strategies have been used, e.g., RL-based methods~\cite{zoph2016neural,pham2018efficient,guo2020}, Evolutionary methods~\cite{liu2017hierarchical,real2019regularized},
gradient-based method~\cite{darts,lian2019towards}, Monte Carlo Tree Search (MCTS) method~\cite{negrinho2017deeparchitect}, etc.

A promising direction to improve the sample efficiency of NAS is to utilize a performance predictor to sample new architectures, a.k.a. \textit{predictor-based NAS}. An early study~\cite{liu2018progressive} trains a surrogate model 
(predictor) to identify promising architectures with increasing complexity. NASBot~\cite{kandasamy2018bayesian} design a distance metric in the architecture space and exploits gaussian process to get the posterior of the architecture performances. Then, it samples new architectures based on the acquisition function calculated using the posterior.
NAO~\cite{nao2018} trains an LSTM-based autoencoder together with a performance predictor based on the latent representation. After updating the latent representation following the predictor's gradients, NAO decodes the latent representation to sample new architectures.

\subsection{Neural Architecture Encoders}

Existing neural architecture encoding schemes include the sequence-based and the graph-based schemes.
In the sequence based scheme, the neural architecture is {\it flattened} into a string encoding the architecture decisions, then encoded using either an LSTM~\cite{nao2018,liu2018progressive,wang2018alphax} or a Multi-Layer Perceptron (MLP)~\cite{liu2018progressive,wang2018alphax}. 
In these methods, the topological information could only be modeled implicitly, which deteriorates the encoder's representational ability. 
Also, the search efficiency would deteriorate since these encoders could not guarantee to map isomorphic architectures~\cite{ying2019bench,stagge2000neural} to the same representation, and data augmentation and regularization tricks are utilized to alleviate this issue~\cite{nao2018}.

Recently, the graph-based encoding scheme that utilizes the topological information explicitly has been used to get better performance. In these graph-based schemes,
graph convolutional networks (GCN)~\cite{kipf2016semi} are usually used to embed the graphs to fixed-length vector representations. For the ``operation on node'' search spaces, in which the operations (e.g., \texttt{Conv3x3}) are on the nodes of the DAG,
GCN can be directly applied~\cite{shi2019multi} to encode architectures, i.e., using adjacency matrix and operation embedding of each node as the input. 
However, for the ``operation on edge'' search spaces, 
in which the operations are on the edges, GCN cannot be applied directly.
A recent study~\cite{guo2019nat} proposes an ad-hoc solution for the ENAS search space. They represent each node by the concatenation of the operation embeddings on the input edges. This solution is contrived and cannot generalized to search spaces where nodes could have different input degrees. Moreover, since the concatenation is not commutative, this encoding scheme could not handle isomorphic architectures correctly.
In brief, existing graph-based encoding schemes are specific to different search spaces, 
and a generic approach for encoding the neural architectures is desirable in the literature.

\section{Method}
\label{sec:napp}

\subsection{Predictor-Based Neural Architecture Search}
\label{sec:pb-nas}

\begin{figure}[tb]
  \begin{center}
    \includegraphics[width=0.98\linewidth]{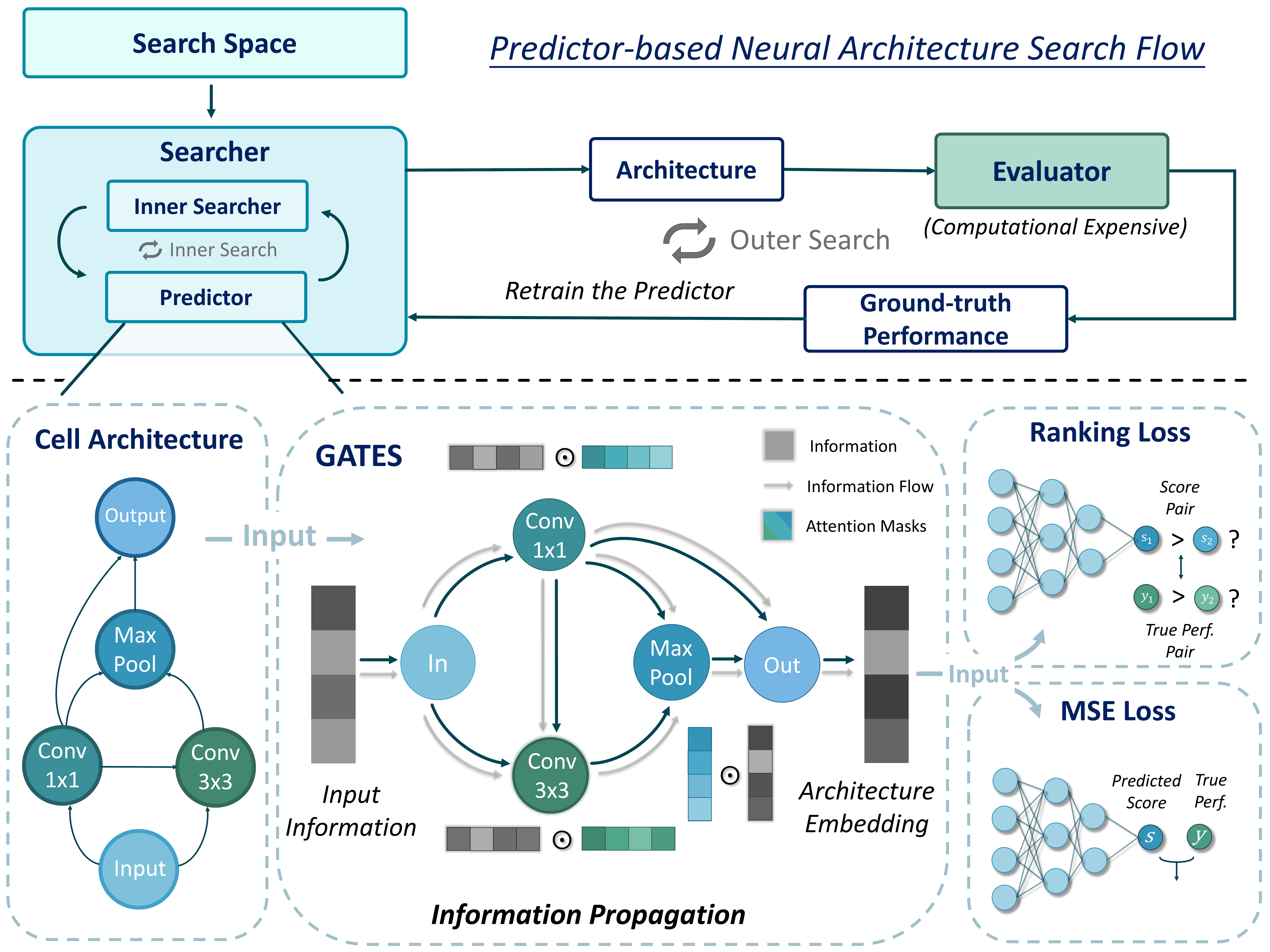}
    \caption{The overview of the proposed algorithm. Upper: The general flow of the predictor-based NAS. Lower: Illustration of the encoding processes of GATES of an OON cell architecture}
    \label{fig:gates}
  \end{center}

\end{figure}

The principle of predictor-based NAS is to increase the sample efficiency of the NAS process, by utilizing an approximated performance predictor to sample architectures that are more worth evaluating.
Generally speaking, the flow of predictor-based NAS could be summarized as in Alg.~\ref{alg:pred-based} and Fig.~\ref{fig:gates}.

\begin{algorithm}[bt]
\begin{algorithmic}[1]
\STATE $\mathcal{A}$: Architecture search space
\STATE $\mbox{P}: \mathcal{A} \to \mathbb{R}$: Performance predictor that outputs the predicted performance given the architecture
\STATE $N^{(k)}$: Number of architectures to sample in the $k$-th iteration
\item[]
\STATE k = 1
\WHILE{$k \leq$ MAX\_ITER}
\STATE Sample a subset of architectures $S^{(k)} = \{a_j^{(k)}\}_{j=1,\cdots,N^{(k)}}$ from $\mathcal{A}$, utilizing $\mbox{P}$
\STATE Evaluate architectures in $S^{(k)}$, get $\tilde{S}^{(k)} =  \{(a_j^{(k)}, y_j^{(k)})\}_{j=1,\cdots,N^{(k)}}$ ($y$ is the performance)
\STATE Optimizing $\mbox{P}$ using the ground-truth architecture evaluation data $\tilde{S} = \cup^{k}_{i=1} \tilde{S}^{(i)}$
\ENDWHILE
\STATE Output $a_{j^*} \in \cup^{k}_{i=1} S^{(i)}$ with best corresponding $y_{j^*}$; Or, $a^* = \mbox{argmax}_{a \in \mathcal{A}} \mbox{P}(a)$
\end{algorithmic}
\caption{The flow of predictor-based neural architecture search}
\label{alg:pred-based}
\end{algorithm}

In line 6 of Alg.~\ref{alg:pred-based}, the architecture candidates are sampled based on the approximated evaluation of the 
predictor.
Utilizing a more accurate predictor, we could choose better architectures for further evaluation.
The better the generalization ability of the predictor is, the fewer architectures are needed to be exactly evaluated to get a highly accurate
predictor. Therefore, the generalization ability of the predictor is crucial for the efficiency and effectiveness of the NAS method.

The model design (i.e., how to encode the neural architectures) of the predictor is crucial to its generalization ability.
We'll introduce our main effort to improve the predictor from the ``model design'' aspect in the following section.

\subsection{GATES: A Generic Neural Architecture Encoder}
\label{sec:method-gates}

\begin{figure*}[tb]
\includegraphics[width=1.0\linewidth]{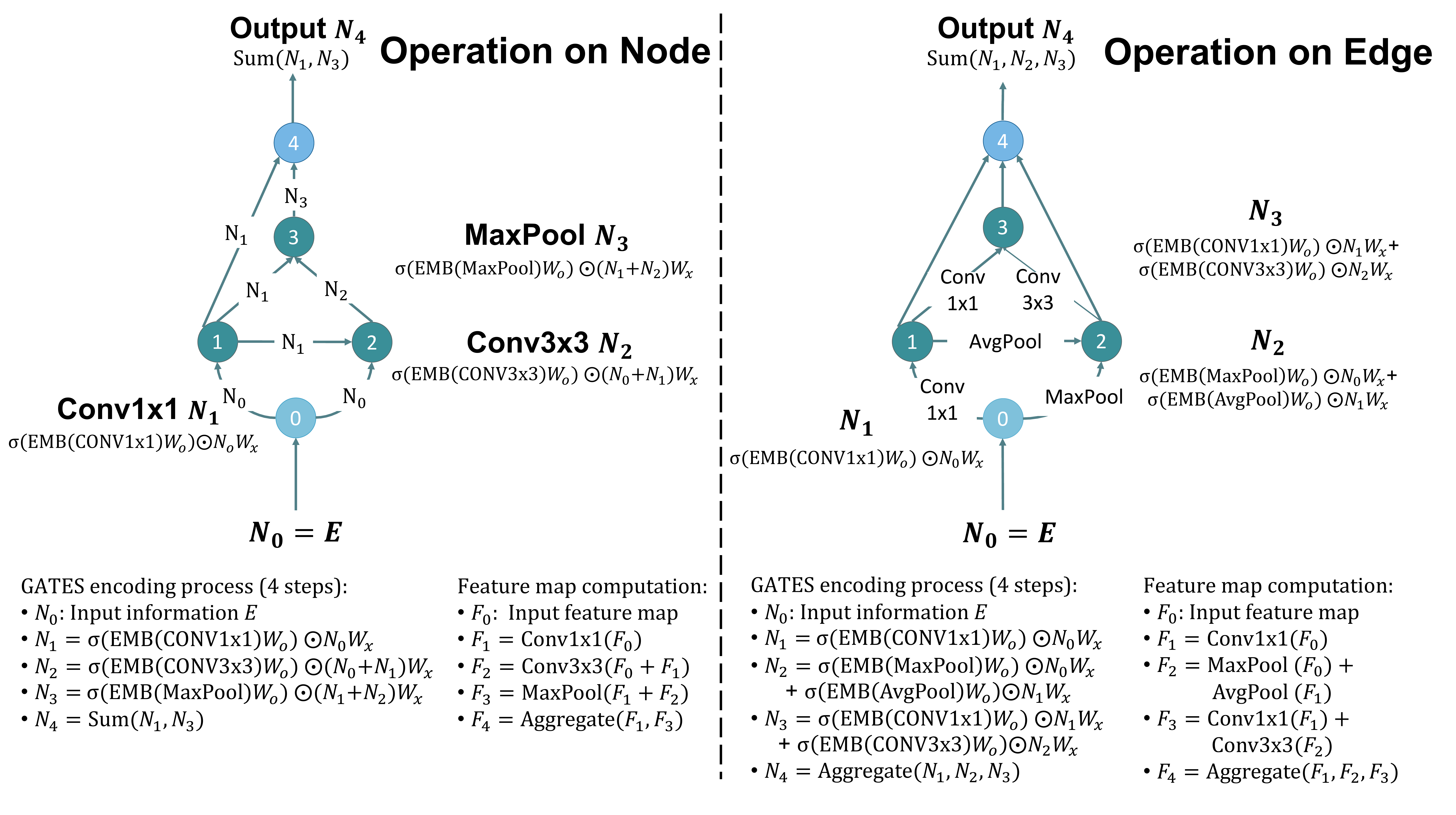}
\caption{Feature map ($F_i$) computation and GATES encoding process ($N_i$). Left: The ``operation on node'' cell search space, where operations (e.g., \texttt{Conv3x3}) are on the nodes of the DAG (e.g., NAS-Bench-101~\cite{ying2019bench}, randomly wired search space~\cite{xie2019exploring}). Right: The ``operation on edge'' cell search space, where operations are on the edges of the DAG. (e.g., NAS-Bench-201~\cite{Dong2020NAS-Bench-201}, ENAS~\cite{pham2018efficient})}
\label{fig:two_search_space}
\end{figure*}

\begin{table*}[b]
  \caption{Notations of GATES. $E$, $\mbox{EMB}$, $W_o$ and $W_x$ are all trainable parameters}
  \label{tab:notation}
  \begin{center}
    \begin{tabular}{cp{8cm}}
      \toprule
      \multirow{2}{*}{$n_i$} & number of input nodes: 1, 1, 2 for NAS-Bench-101, NAS-Bench-201 and ENAS, respectively \\
      \specialrule{0em}{1pt}{4pt}
      $N_o$ &  number of operation primitives\\
      $h_o$ & embedding size of operation\\
      $h_i$ & embedding size of information\\\cmidrule(lr){1-2}
      $E \in \mathbb{R}^{n_i\times h_i}$  & the embedding of the information at the input nodes\\
      $\mbox{EMB} \in \mathbb{R}^{N_o \times h_o}$ & the operation embeddings \\
      $W_o \in \mathbb{R}^{h_o \times h_i}$ & the transformation matrix on the operation embedding\\
      $W_x \in \mathbb{R}^{h_i \times h_i}$ & the transformation matrix on the information\\\bottomrule
    \end{tabular}
  \end{center}
\end{table*}

A performance predictor $\mbox{P}$ is a model that takes a neural architecture $a$ as input, and outputs a predicted score $\hat{s}$. Usually, the performance predictor is constructed by an encoder followed by an MLP, as shown in Eq.~\ref{eqn:pred}. 
The encoder $\mbox{Enc}$ maps a neural architecture into a continuous embedding space, and its design is vital to the generalization ability of the performance predictor. Existing encoders include the sequence-based ones (e.g., MLP, LSTM) and the graph-based ones (e.g., GCN). We design a new graph-based neural architecture encoder GATES that is more suitable for modeling neural architectures.

\begin{equation}
    \begin{aligned}
    \hat{s} = \mbox{P}(a) = \mbox{MLP}(\mbox{Enc}(a))
    \end{aligned}
    \label{eqn:pred}
\end{equation}

To encode a cell architecture into an embedding vector, GATES follows the ideology of modeling the information flow in the architecture, and uses the output information as the embedding of the architecture. The notations are summarized in Table~\ref{tab:notation}.

Specifically, we models the input information as the embedding of the input nodes $E \in \mathbb{R}^{n_i\times h_i}$, where $n_i$ is the number of input nodes,
and $h_i$ is the embedding size of the information.
The information (embedding of the input nodes) is then ``processed'' by the operations and ``propagates'' along the DAG.

The encoding process of GATES goes as follows: Upon each unary operation $o$ (e.g., \texttt{Conv3x3}, \texttt{MaxPool}, etc.), the input information $x_{\mbox{in}}$ of this operation is processed by a linear transform $W_x$ and then elementwise multiplied with a soft attention mask $m=\sigma(\mbox{EMB}(o)W_o) \in \mathbb{R}^{1 \times h_i}$.
\begin{equation}
  x_{\mbox{out}} = m \odot x_{\mbox{in}} W_x
\end{equation}
where $\odot$ denotes the elementwise multiplication. And the mask $m$ is calculated from the operation embedding $\mbox{EMB}(o) = \mbox{onehot}(o)^T \mbox{EMB} \in \mathbb{R}^{1 \times h_o}$.

Multiple pieces of information are aggregated at each node using summation.
Finally, after obtaining the virtual information at all the nodes, the information at the output node is used as the embedding of the entire cell architecture. For search spaces with multiple cells (e.g., normal and reduce cells in ENAS), GATES encodes each cell independently, and concatenate the embeddings of cells as the embedding of the architecture.

Fig.~\ref{fig:two_search_space} illustrates two examples of the encoding process in the OON and OOE search spaces. As can be seen, the encoding process of GATES mimics the actual feature map computation.
For example, in the example of the OON search space, the actual feature map computation at node 2 is $F_2 = \mbox{Conv3x3}(F_0 + F_1)$, where $F_i$ is the feature map at node $i$.
To model the information processing of this feature map computation, GATES calculates the information (node embedding) at node 2 by $N_2 = \sigma(\mbox{EMB}(\mbox{Conv3x3})W_o) \odot (N_0 + N_1) W_x$, where $\sigma(\cdot)$ is the sigmoid function, and $W_o \in \mathbb{R}^{h_o \times h_i}$ is a transformation matrix that transforms the $h_o$-dim operation embedding into a $h_i$-dim feature.
That is to say, the summation of feature maps $F_0 + F_1$ corresponds to the summation of the virtual information $N_0 + N_1$, and the data processing function $o(\cdot)$ (\texttt{Conv3x3}) corresponds to a transform
$f(\cdot)$ that processes the information $x = N_0 + N_1$ by $f_o(x) = \sigma(\mbox{EMB}(o)W_o) \odot x W_x$.

Intuitively, to model a cell architecture, GATES models the operations in the architecture as the ``soft gates'' that control the flow of the virtual information, and the output information is used as the embedding of the cell architecture. 
The key difference between GATES and GCN is: In GATES, the operations (e.g., \texttt{Conv3x3}) are modeled as the processing of the node attributes (i.e., virtual information), whereas GCN models them as the node attributes themselves.

The representational power of GATES for neural architectures comes from two aspects: 1)
The more reasonable modeling of the operations in data-processing DAGs. 2) The intrinsic proper handling of DAG isomorphism. 
The discussion and experiments on how GATES handles the isomorphism are in the ``Discussion on Isomorphism'' section in the appendix.

In practice, to calculate the information propagation following the topological order of different graphs in a batched manner, we use a stack of GATES layers. In the forward process of each layer, one step of information propagation is taken place at every node. That is to say, if a graph is input to a GATES encoder with $N$ layers, the information is propagated and aggregated for $N$ steps along the graph. 
The batched formulas and specific implementations of a GATES layer for OON and OOE search spaces are elaborated in the ``Implementation of GATES'' section in the appendix.

\subsubsection{The Optimization of GATES}

The most common practice~\cite{liu2018progressive,nao2018} to train the architecture performance predictors is to minimize the Mean Squared Error (MSE) between the predictor outputs and the true performances.
\begin{equation}
  L(\{a_j, y_j\}_{j=1,\cdots,N}) = \sum_{j=1}^N(\mbox{P}(a_j) - y_j)^2
\end{equation}
where $a_j$ denotes one architecture, and $y_j$ denotes the true performance of $a_j$.

In NAS applications, what is really required to guide the search of architectures is the relative ranking order of architectures rather than the absolute performance values. In this paper, we adopt Kendall's Tau ranking correlation~\cite{sen1968estimates} as the measure as the direct criterion for evaluating architecture predictors. And since ranking losses are better surrogate losses~\cite{chen2009,liu2009learning,xu2019renas} for the ranking correlation than the regression loss, 
in addition to the MSE loss,
we use a hinge pair-wise ranking loss with margin $m$=0.1 to train the predictors.\footnote{A more comprehensive comparison of the MSE regression loss and multiple ranking losses is shown in the appendix.}

\begin{equation}
  L(\{a_j, y_j\}_{j=1,\cdots,N}) = \sum_{j=1}^N\sum_{i, y_i > y_j} \max[0, m - (\mbox{P}(a_i) - \mbox{P}(a_j))]
\end{equation}

\subsection{Neural Architecture Search Utilizing the Predictor}
\label{sec:method-nas}

We follow the flow in Alg.~\ref{alg:pred-based} to conduct the architecture search. There are multiple ways of utilizing the predictor $\mbox{P}$ to sample architectures (line 6 in Alg.~\ref{alg:pred-based}), i.e., the choice of the inner search method. In this work, we use two inner search methods for sampling architecture for further evaluation:\footnote{Note that this inner search component could be easily substituted with other search strategies.}
\begin{itemize}
    \item Random sample $n$ architectures from the search space, then choose the best $k$ among them according to the evaluation of the predictor.

    \item Search with Evolutionary Algorithm (EA) for $n$ steps, and then choose the best $k$ with the highest predicted scores among the seen architectures.

\end{itemize}

Compared with the evaluation (line 7 in Alg.~\ref{alg:pred-based}) in the outer search process, the evaluation of each architecture in the inner search process is very efficient with only a forward pass of the predictor.
The sample ratio $r=\frac{n}{k}$ indicates the equivalent number of the architectures need to be evaluated by the predictor to make one sample decision. And it is not the case that bigger $r$ leads to better sample efficiency of the overall NAS process. If $n$ is too large (the limiting case is to exhaustive test the whole search space with $n=|\mathcal{A}|$), the sampling process would overfit onto exploiting the current performance predictor and fails to explore. Therefore, there is a trade-off between exploration and exploitation controlled by $n$, which we verify in Sec.~\ref{sec:exp-nas-nb101}.

\section{Experiments}
\label{sec:exp}
The experiments in Sec.~\ref{sec:exp-nasbench} and Sec.~\ref{sec:exp-nasbench201} verify the effectiveness of the GATES encoder on both the OON and OOE search spaces. 
Then, in Sec.~\ref{sec:exp-nas-nb101}, we demonstrate that by utilizing GATES, the sample efficiency of the NAS process surpasses other searching strategies, including the predictor-based methods with other baseline encoders.
Finally, in Sec.~\ref{sec:exp-nas-enas}, we apply the proposed algorithm to the ENAS search space. 

\subsection{Predictor Evaluation on NAS-Bench-101}
\label{sec:exp-nasbench}

\subsubsection{Setup}
NAS-Bench-101~\cite{ying2019bench} provides the performances of the 423k unique architectures in a search space. The NAS-Bench-101 search space is an OON search space, in which sequence based encoding schemes~\cite{wang2018alphax}, and graph based encoding schemes~\cite{shi2019multi} are proposed for encoding architectures. We use the Kendall's Tau ranking correlation~\cite{sen1968estimates} as the measure for evaluating the architecture performance predictors. 
The first 90\% (381262) architectures are used as the training data, and the other 42362 architectures are used for testing.\footnote{See ``Setup and Additional Results'' section in the appendix for more details.}

We conduct a more comprehensive comparison of the MSE loss and multiple ranking losses on NAS-Bench-101, and the results are shown in the appendix. We find that compared to the MSE loss, ranking losses bring consistent improvements, and hinge pair wise loss is a good choice. Therefore, in our experiments, unless otherwise stated, the hinge pairwise loss with margin 0.1 is used to train all the predictors.

\subsubsection{Results}
\label{sec:exp-nasnbech-gates}

\begin{table*}[tb]
  \caption{The Kendall’s Tau of using different encoders on the NAS-Bench-101 dataset. The first 90\% (381262) architectures in the dataset are used as the training data, and the other 42362 architectures are used as the testing data}
  
\label{table:gates-nb101}
\begin{center}
\resizebox{1.0\textwidth}{!}{
\begin{tabular}{ccccccccc}
\toprule
\multirow{2}{*}{Encoder} & \multicolumn{8}{c}{Proportions of 381262 training samples}\\ 
\cmidrule(lr){2-9} & 0.05\% & 0.1\% & 0.5\% & 1\% & 5\% & 10\% & 50\% & 100\% \\ \midrule
MLP~\cite{wang2018alphax}     &    0.3971    &    0.5272   &    0.6463   &   0.7312  & 0.8592    & 0.8718     &     0.8893     &  0.8955  \\
LSTM~\cite{wang2018alphax}     &      0.5509  &    0.5993  &   0.7112    & 0.7747     &0.8440     &0.8576      &    0.8859     & 0.8931       \\
GCN (w.o. global node) &    0.3992    &   0.4628    &  0.6963     &    0.8243 &   0.8626  &    0.8721  &    0.8910 &   0.8952    \\
GCN (global node)~\cite{shi2019multi} & 0.5343 & 0.5790 & 0.7915 & 0.8277 &  0.8641 & 0.8747 & 0.8918 &  0.8950\\
\hline
GATES & {\bf 0.7634} & {\bf 0.7789} & {\bf 0.8434} & {\bf 0.8594} & {\bf 0.8841} & {\bf 0.8922} & {\bf  0.9001} & {\bf 0.9030}\\\bottomrule
\end{tabular}
}
\end{center}
\end{table*}

\begin{table*}[th]
    \caption{N@K on NAS-Bench-101. All predictors are trained with 0.1\% of the training data (i.e., 381 architectures)}
    \label{table:natk-nb101}
    \begin{center}
    \begin{tabular}{c@{\hskip 0.02\linewidth}cccc}
    \toprule
    \multirow{2}{*}{Encoder} & \multicolumn{2}{c}{Ranking Loss} &  \multicolumn{2}{c}{Regression Loss} \\ 
    \cmidrule(lr){2-3}\cmidrule(lr){4-5} & N@5 & N@10 & N@5 & N@10  \\ \midrule
    MLP~\cite{wang2018alphax}   &    57 (0.13\%)    &    58 (0.13\%)   &    1397 (3.30\%)   &   552 (1.30\%)    \\
    LSTM~\cite{wang2018alphax}     &      1715 (4.05\%)  &   1715 (4.05\%) &   1080 (2.54\%)   & 312 (0.73\%)  \\
    GCN ~\cite{shi2019multi} & 2025 (4.77\%) & 1362 (3.21\%) & 405 (0.95\%) & 405 (0.95\%) \\
    \hline
    GATES & {\bf 22 (0.05\%)} & {\bf 22 (0.05\%)} & {\bf 27 (0.05\%)} & {\bf 27 (0.05\%)} \\\bottomrule
    \end{tabular}
  \end{center}
\end{table*}

Table~\ref{table:gates-nb101} shows the comparison of the GATES encoder and various baseline encoders trained using different proportions of the training data. 
As can be seen, GATES could achieve higher Kendall's Taus on the testing architectures than the baseline encoders consistently with different training proportions. The advantages are especially significant when there are few training architectures. For example, when only 190 (0.05\%) architectures are seen by the performance predictor, utilizing the same training settings, GATES achieves a test Kendall's Tau of \textit{0.7634},
whereas the Kendall's Tau results achieved by MLP, LSTM, and the best GCN variant are 0.3971, 0.5509 and 0.5343, respectively. This demonstrates the surpassing generalization ability of the GATES encoder, which enables one to learn a good performance predictor for unseen architectures after evaluating only a small set of architectures.

\begin{figure*}[tb]
  \begin{center}
    \subfigure[NAS-Bench-101]{
      \includegraphics[width=0.45\linewidth]{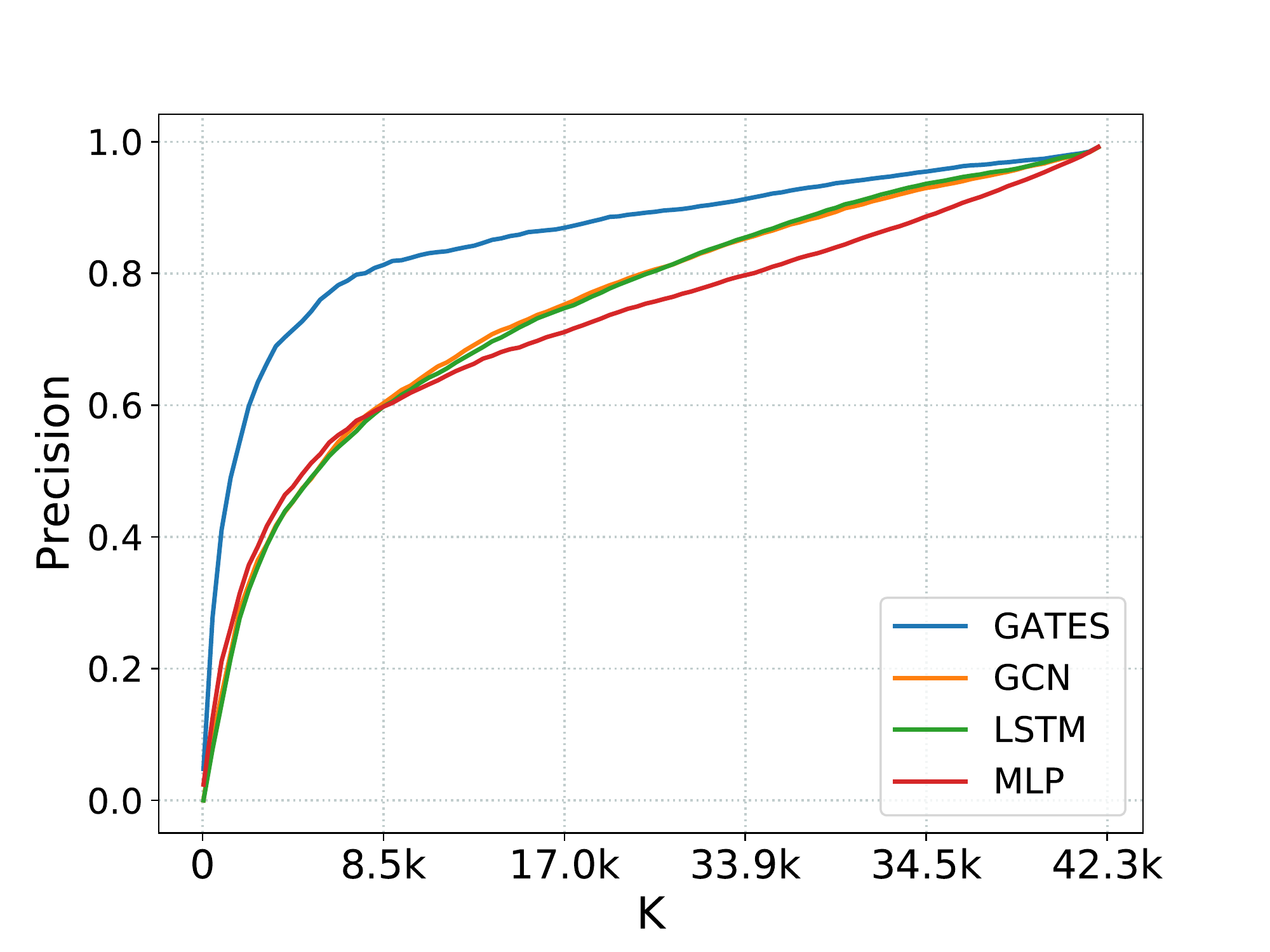} 
      \label{fig:patk_nb101}
    }
    \subfigure[NAS-Bench-201]{
      \includegraphics[width=0.45\linewidth]{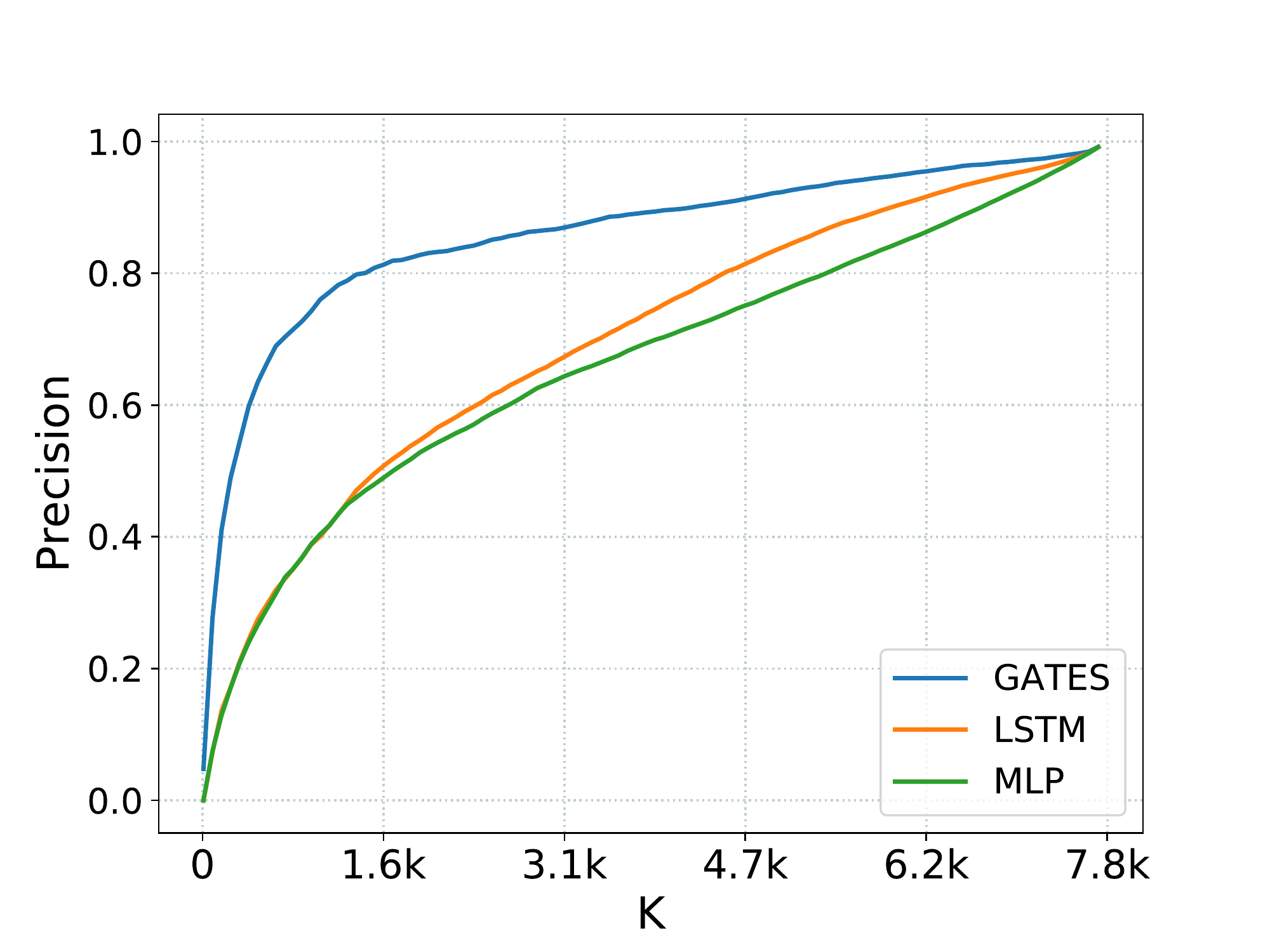} 
      \label{fig:patk_nb201}
    }
    \caption{Precision@K}
  \end{center}
  \label{fig:patk}
\end{figure*}

In the Kendall's Tau measure, all discordant pairs are treated equally. However, in NAS applications, the relative rankings among the poorly performing architectures are not of concern. Therefore, we compare different predictors in the form of other measures that have a more direct correspondence with the NAS flow: 1) N@K: The best true ranking among the top-K architectures selected according to the predicted scores. 2) Precision@K: The proportion of true top-K architectures among the top-K predicted architectures. Table.~\ref{table:natk-nb101} and Figure.~\ref{fig:patk_nb101} show these two measures of the predictors with different encoders on the testing set of NAS-Bench-101. As can be seen, GATES achieves consistently better performances than other encoders across different Ks.

\subsection{Predictor Evaluation on NAS-Bench-201}
\label{sec:exp-nasbench201}

\subsubsection{Setup}
NAS-Bench-201~\cite{Dong2020NAS-Bench-201} is another NAS benchmark that provides the performances of 
15625 architectures in an OOE search space. In our experiments, we use the first 50\% (7813) as the training data, and the remaining 7812 architectures as the testing data.
Since GCN encoders could not be directly applied to the OOE search spaces, 
we compare GATES with the sequence-based encoders: MLP and LSTM.\footnote{We also implement an ad-hoc solution of applying GCN on OOE architectures referred to as the Line Graph GCN solution, in which the graph is first converted to a line graph. See ``Setup and Additional Results'' section in the appendix for more details.} 

\addtolength{\tabcolsep}{1pt}
\begin{table*}[tb]
\caption{The Kendall’s Tau of using different encoders on the NAS-Bench-201 dataset. The first 50\% (7813) architectures in the dataset are used as the training data, and the other 7812 architectures are used as the testing data}
\label{table:gates-nb201}
\begin{center}
\begin{tabular}{cccccc}
\toprule
\multirow{2}{*}{Encoder} & \multicolumn{5}{c}{Proportions of 7813 training samples}\\ 
\cmidrule(lr){2-6} & 1\% & 5\% & 10\% & 50\% & 100\% \\\midrule
MLP~\cite{wang2018alphax}   &  0.0974 & 0.3959 & 0.5388 & 0.8229 & 0.8703\\
  LSTM~\cite{wang2018alphax}  & 0.5550 & 0.6407 & 0.7268 & 0.8791 & 0.9002\\
\hline
GATES & {\bf 0.7401} & {\bf 0.8628} & {\bf 0.8802} & {\bf 0.9192} & {\bf 0.9259}\\\bottomrule
\end{tabular}
\end{center}
\end{table*}
\addtolength{\tabcolsep}{-1pt}

\begin{table}[th]
  \caption{N@K on NAS-Bench-201. All the predictors are trained using 10\% of the training data (i.e., 781 architectures)}
  \label{table:natk-nb201}
  \begin{center}
    \begin{tabular}{c@{\hskip 0.02\linewidth}cccc}
      \toprule
      \multirow{2}{*}{Encoder} & \multicolumn{2}{c}{Ranking Loss} &  \multicolumn{2}{c}{Regression Loss} \\ 
      \cmidrule(lr){2-3}\cmidrule(lr){4-5} & N@5 & N@10 & N@5 & N@10  \\ \midrule
      MLP~\cite{wang2018alphax}  &    7 (0.09\%)    &    7 (0.09\%)   &   1538 (19.7\%)   &   224 (3.87\%)    \\
      LSTM~\cite{wang2018alphax}     &  8 (1.02\%)  &   2 (0.01\%) &  250 (6.65\%)   & 234 (2.99\%)  \\
      \hline
      GATES & {\bf 1 (0.00\%)} & {\bf 1 (0.00\%)} & {\bf 1 (0.00\%)} & {\bf 1 (0.00\%)} \\\bottomrule
    \end{tabular}
  \end{center}
\end{table}

\subsubsection{Results}
Table~\ref{table:gates-nb101} shows the evaluation results of GATES. GATES could achieve significantly higher ranking correlations than the baseline encoders, especially when there are only a few training samples. For example, with 78 training samples, ``GATES + Pairwise loss'' could achieve a Kendall's Tau of 0.7401, while the best baseline result is 0.5550 (``LSTM + Pairwise loss'').

The N@K and Precision@K measures on NAS-Bench-201 are shown in Table~\ref{table:natk-nb201} and Fig.~\ref{fig:patk_nb201}, respectively. 
We can see that GATES can achieve an N@5 of 1 on the 7812 testing architectures, with either ranking loss or regression loss. And, not surprisingly, GATES outperforms the baselines consistently on the Precision@K measure too.

\subsection{Neural Architecture Search on NAS-Bench-101}
\label{sec:exp-nas-nb101}

Equipped with a better performance predictor, the sample efficiency of the predictor-based NAS process can be significantly improved. To verify that, we conduct the architecture search on NAS-Bench-101 using various searching strategies. As the baseline of our method, 
we run a random search, regularized evolution~\cite{real2019regularized},
and predictor-based NAS methods equipped with the baseline encoders (i.e., LSTM, MLP, GCN).

\begin{figure*}[bt]
  \subfigure[RS inner search method ($r=500$)]{
    \includegraphics[width=0.45\linewidth]{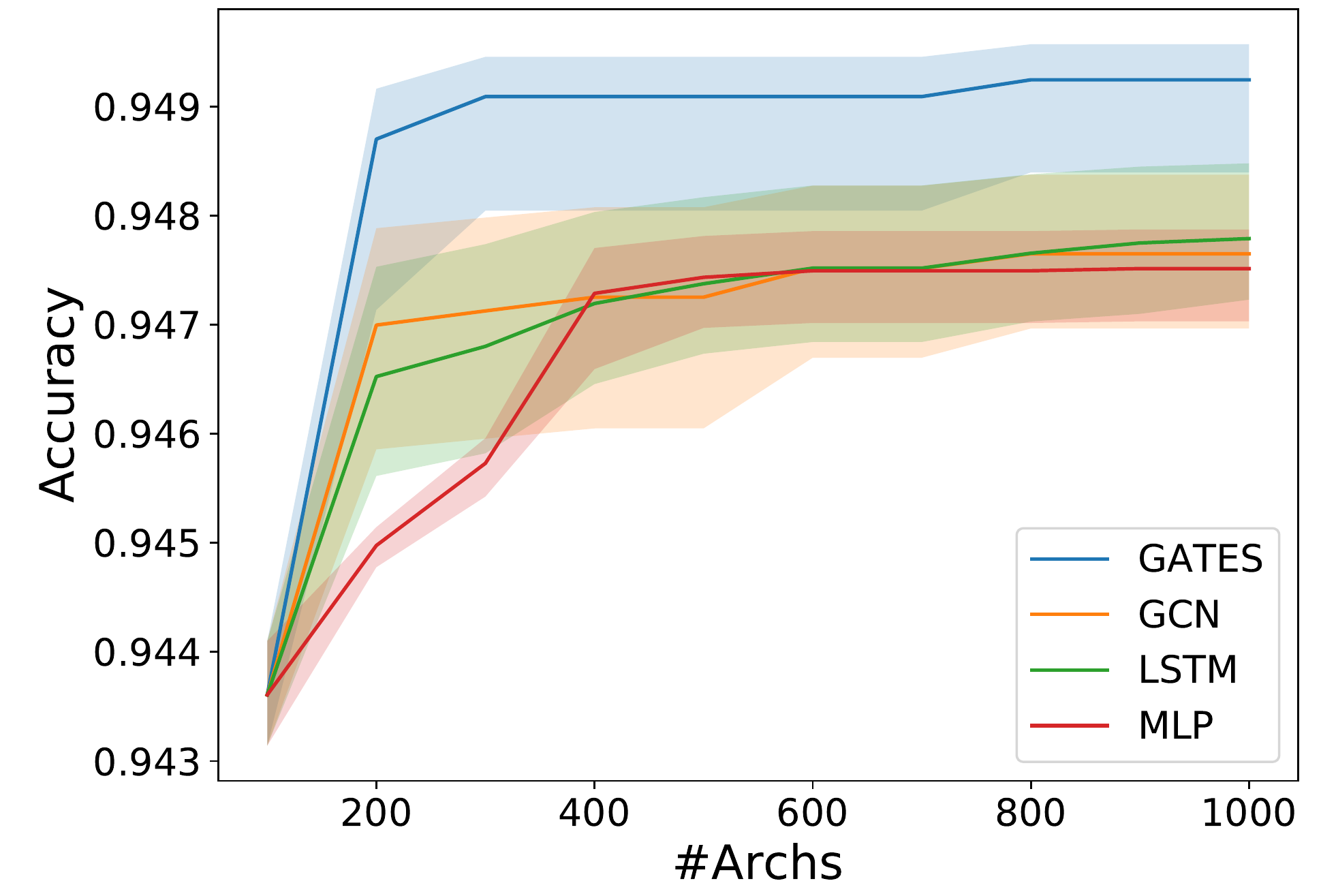} 
    \label{fig:acc_line_a}
  }
  \subfigure[EA inner search method ($r=100$)]{
    \includegraphics[width=0.45\linewidth]{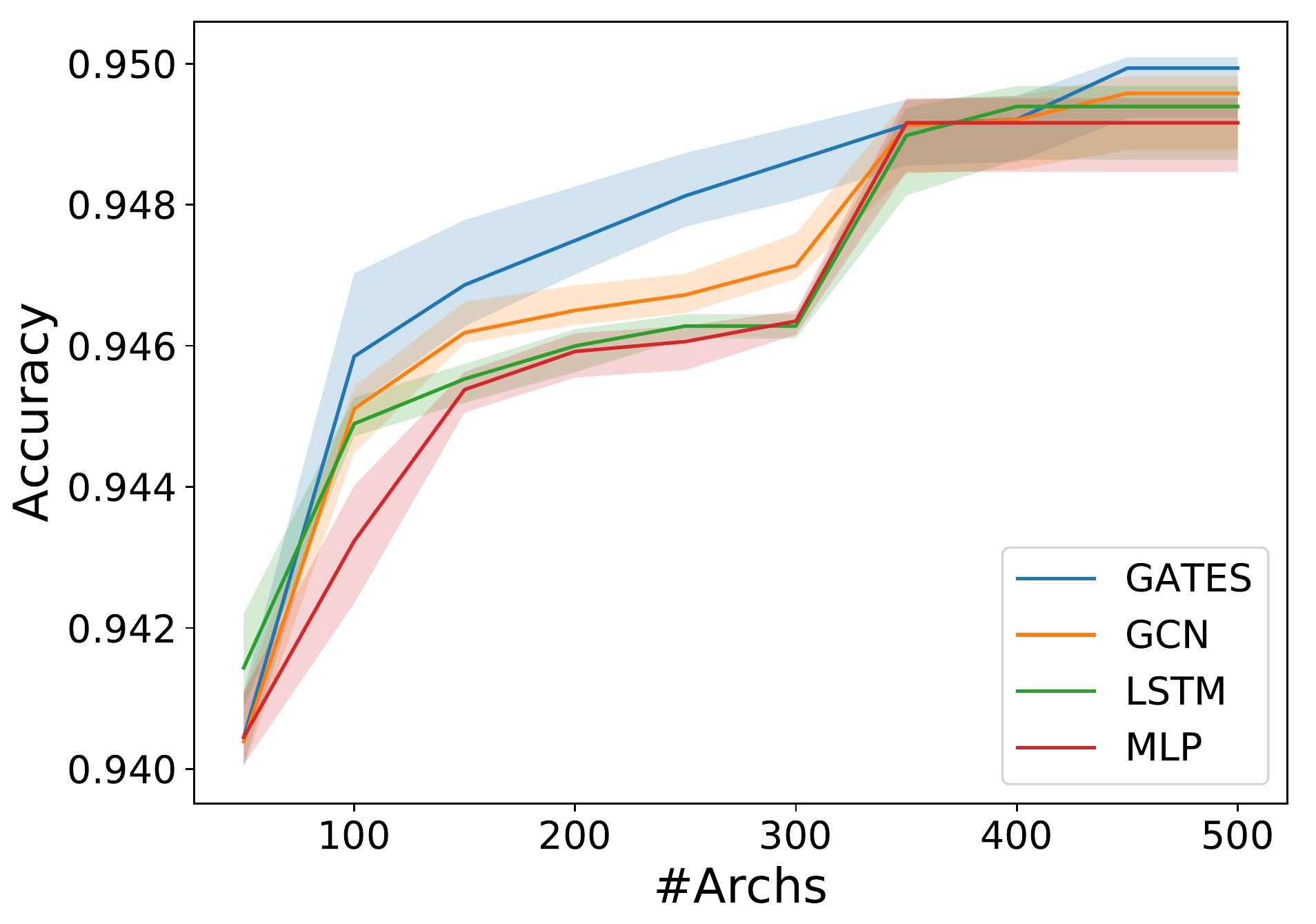} 
    \label{fig:acc_line_b}
  }
\caption{Comparison of predictor-based NAS with different encoders: The best validation accuracy during the search process over 10/15 runs for the RS and EA inner serach method, respectively. $r$ is the sample ratio (see Sec.~\ref{sec:method-nas})}
\label{fig:accs_under_archs}
\end{figure*}

\begin{figure*}[tb]
  \subfigure[Comparison of search methods]{
    \includegraphics[width=0.47\linewidth]{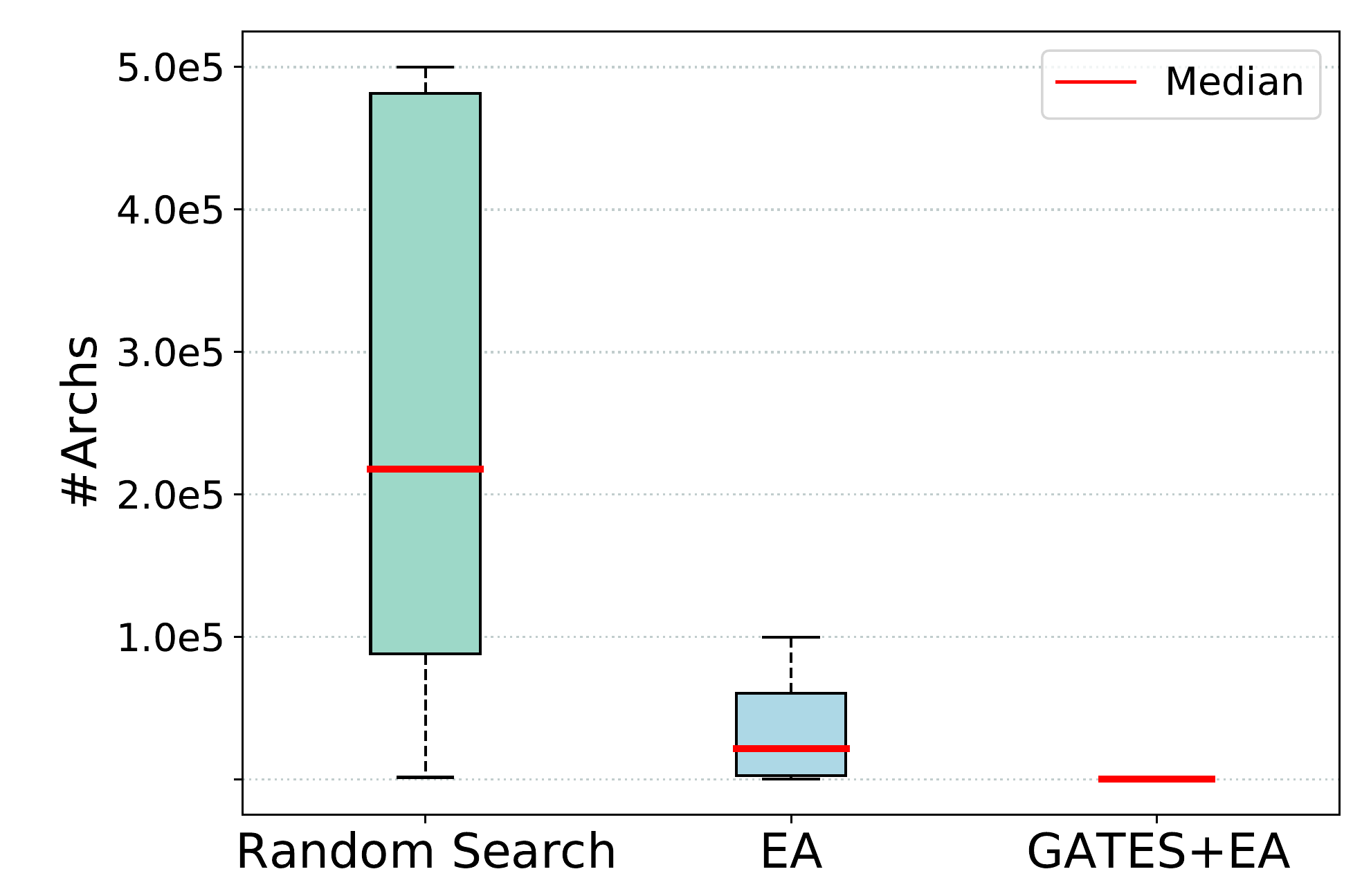} 
    \label{fig:box-nb101}
  }
  \subfigure[Ablation study of the sample ratio $r$]{
    \includegraphics[width=0.45\linewidth]{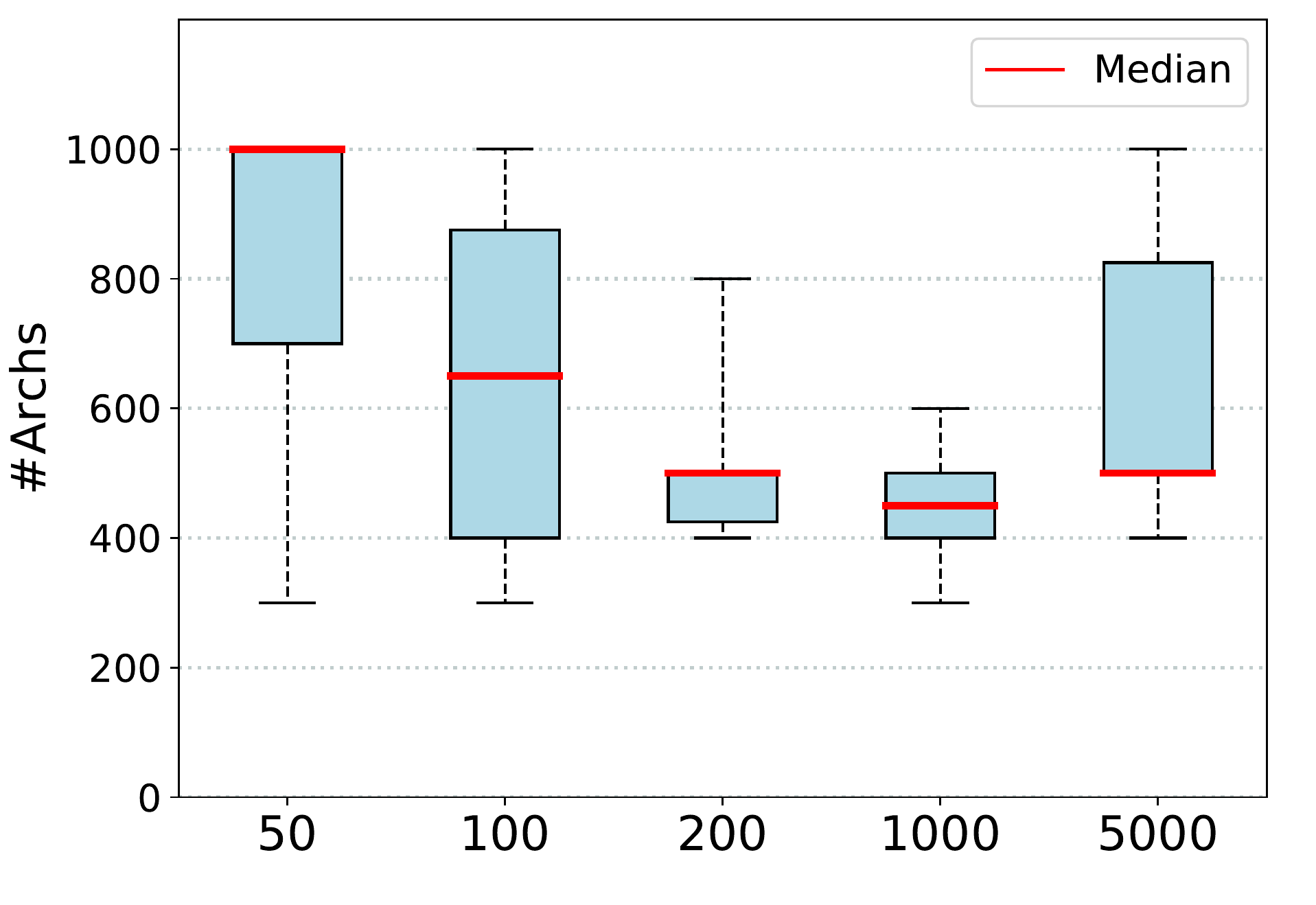} 
    \label{fig:sample_ratio}
  }
\caption{Left: Number of architectures evaluated to acquire the best validation accuracy on NAS-Bench-101 over 100 runs. We use the mean validation accuracy as the search reward. GATES-powered predictor-based NAS is 511.0$\times$ and 59.25$\times$ more sample efficient than random search and regularized evolution. Right: 
  Number of architectures 
  evaluated to acquire the best validation accuracy over 10 runs with different $r$}
\label{fig:box_ablation}
\end{figure*}

\subsubsection{Comparison of Sample Efficiency}
The results of running predictor-based NAS methods with different encoders are shown in Fig.~\ref{fig:accs_under_archs}. We conduct experiments with two inner search methods: random search, and evolutionary algorithm. 
In each stage, 100 random samples are used to train the predictor (50 for evolutionary algorithm), and the predictor is trained for 50 epochs with hinge ranking loss. When using random search, $n=2500$ architectures are randomly sampled, and the top $k=5$ architectures with high predicted scores are chosen to be further evaluated by the ground truth evaluator. When using the evolutionary algorithm for the inner search, $n$ is set to 100, and $k$ is set to $1$. And the population and tournament size is 20 and 5, respectively. 
We can see that the sample efficiency using GATES surpasses the baselines with different inner search methods. This verifies the analysis that utilizing a better neural architecture encoder in the predictor-based NAS flow leads to better sample efficiency.

The comparison of the sample efficiency of two baseline searching strategies and the predictor-based method with GATES is shown in Fig.~\ref{fig:box-nb101}.
The median counts of evaluated architectures of RS, Regularized EA and GATES-powered NAS over 100 runs are 220400, 23700 and 400 (50 as the granularity), respectively.
GATES-powered NAS is 551.0$\times$ and 59.25$\times$ more sample efficient than the random search and evolution algorithm.

\subsubsection{Ablation Study of the Sample Ratio $r$}
The ablation study of the sample ratio $r$ (Sec.~\ref{sec:method-nas}) is shown in Fig.~\ref{fig:sample_ratio}.
We run GATES-powered predictor-based search with evolutionary algorithm, and shows the architectures needed to evaluate before finding the architecture with the best validation accuracy. 
We can see that the sample ratio $r$ should be neither too big nor too small, since a too small $n$ leads to bad exploitation and a too large $n$ leads to bad exploration.

\subsection{Neural Architecture Search in the ENAS Search Space}
\label{sec:exp-nas-enas}

\addtolength{\tabcolsep}{1pt}
\begin{table}[tb]
  \caption{Comparison of NAS-discovered architectures on CIFAR-10}
  \label{table:nas-enas}
  \begin{center}
    \begin{tabular}{cccc}
      \toprule
      Method & Test Error (\%) & \#Params (M) & \#Archs Evaluated\\\midrule
      NASNet-A + cutout~\cite{zoph2016neural}         & 2.65 & 3.3   & 20000 \\
      AmoebaNet-B + cutout~\cite{real2019regularized} & 2.55 & 2.8   & 27000 \\
      NAONet~\cite{nao2018}                           & 2.98 & 28.6  & 1000  \\
      PNAS~\cite{liu2018progressive}                  & 3.41 & 3.2   & 1160  \\\cmidrule(lr){1-4}
      NAONet-WS$^\dagger$~\cite{nao2018}                        & 3.53 & 2.5   & -     \\
      DARTS+cutout$^\dagger$~\cite{darts}                       & 2.76 & 3.3   & -     \\
      ENAS + cutout$^\dagger$~\cite{pham2018efficient}          & 2.89 & 4.6   & -     \\ \hline
      Ours + cutout                                   & 2.58 & 4.1  & 800   \\ \bottomrule
    \end{tabular}
      \begin{minipage}{1.0\textwidth}
      $\dagger$: As discussed in Sec.~\ref{sec:related_work}, the challenge faced by one-shot NAS lies in the evaluation correlation rather than sample efficiency, thus we do not report the sample efficiency of the one-shot (parameter sharing) NAS methods.
    \end{minipage}
  \end{center}
\end{table}
\addtolength{\tabcolsep}{-1pt}

In this section, we apply our method on the ENAS search space. This search space is an OOE search space that is much larger than the benchmark search spaces. We first randomly sample 600 architectures and train them for 80 epochs.
Then we train a GATES predictor using the performance of the 600 architectures and use it
to sample 200 architectures, by randomly sampling 10k architectures and taking the top 200
with the highest predicted scores (sample ratio $r=50$). After training these 200 architectures for 80 epochs, we pick the architecture with the best validation accuracy. Finally, after the channel and layer augmentation, the architecture is trained from scratch for 600 epochs.

The comparison of the test errors of different architectures is shown in Table~\ref{table:nas-enas}, and the discovered architecture is shown in the appendix. As can be seen, our discovered architecture can achieve a
test error rate of 2.58\%, which is better than those architectures discovered with parameter sharing evaluation. Compared to the other methods, much fewer samples are truly evaluated to discover an architecture with better or comparable performance. When transferred to ImageNet, the discovered architecture achieves a competitive top-1 error of 24.1\% with 5.6M parameters.

\section{Conclusion}
\label{sec:conclusion}
In this paper, we propose GATES, a graph-based neural architecture encoder with better representation ability for neural architectures. Due to its reasonable modeling of the neural architectures and intrinsic ability to handle DAG isomorphism, GATES significantly improves the architecture performance predictor for different cell-based search spaces. Utilizing GATES in the predictor-based NAS flow leads to consistent improvements in sample efficiency. Extensive experiments demonstrate the effectiveness and rationality of GATES. Employing GATES to encode architectures in larger or hierarchical topological search spaces is an interesting future direction.

\section*{Acknowledgments}
This work was supported by National Natural Science Foundation of China (No. 61832007, 61622403, 61621091, U19B2019), Beijing National Research Center for Information Science and Technology (BNRist). The authors thank Novauto for the support.

\clearpage

%
%
\bibliographystyle{splncs04}
\bibliography{egbib}

\pagestyle{headings}

\title{Appendices: A Generic Graph-based Neural Architecture Encoding Scheme for Predictor-based NAS}

\author{}
\institute{}
\email{}

\titlerunning{Appendices of GATES}
\authorrunning{X. Ning et al.}

\maketitle

\section{Implementation of GATES}
\begin{table*}[tb]
\caption{Notations used in the batched computation of the GATES encoder}
\label{tab:notation}
\begin{center}
\begin{tabular}{cp{8cm}}
\toprule
\multirow{2}{*}{$V$} & maximum number of nodes: 7, 4, 6 for NAS-Bench-101~\cite{ying2019bench}, NAS-Bench-201~\cite{Dong2020NAS-Bench-201} and ENAS~\cite{pham2018efficient}, respectively \\
\specialrule{0em}{1pt}{4pt}
\multirow{2}{*}{$n_i$} & number of input nodes: 1, 1, 2 for NAS-Bench-101, NAS-Bench-201 and ENAS, respectively \\
\specialrule{0em}{1pt}{4pt}
$N_o$ &  number of operation primitives\\\cmidrule(lr){1-2}
$h_o$ & embedding size of operation\\
$h_i^{(k)}$ & embedding size of information in the $k$-th layer\\
$E \in \mathbb{R}^{n_i\times h^{(0)}_i}$  & the embedding of the information at the input nodes\\
$\mbox{EMB} \in \mathbb{R}^{N_o \times h_o}$ & the operation embeddings \\
\multirow{2}{*}{$W^{(k)}_o\in \mathbb{R}^{h_o \times h_i^{(k)}}$} & the transformation matrix on the operation embedding (the $k$-th layer)\\
\multirow{2}{*}{$W^{(k)}_x \in \mathbb{R}^{h_i^{(k-1)} \times h_i^{(k)}}$} & the transformation matrix on previous layer's output information (the $k$-th layer)\\\cmidrule(lr){1-2}
$b$ & batch size\\
$A \in \mathbb{R}^{b\times V\times V}$ & adjacency matrix\\
$X^{(k)} \in \mathbb{R}^{b \times V \times h_i^{(k)}}$ & the output virtual information of the $k$-th layer\\\hline
\specialrule{0em}{3pt}{3pt}
\multirow{2}{*}{$\mbox{EMB}(o) \in \mathbb{R}^{b \times V \times h_o}$} & (NAS-Bench-101) the embeddings of the operations on nodes\\
\multirow{2}{*}{$\mbox{EMB}(o) \in \mathbb{R}^{b \times V \times V \times h_o}$} & (NAS-Bench-201) the embeddings of the operations on edges\\
$n_d$& (ENAS) maximum input degree of nodes\\
\multirow{2}{*}{$\mbox{EMB}(o_d) \in \mathbb{R}^{b \times V \times V \times h_o}$} & (ENAS) the embeddings of operations on the $d$-th input edge for nodes 
\\\bottomrule
\end{tabular}
\end{center}
\end{table*}

In practice, to calculate the information propagation following the topological order of different graphs in a batched manner, we use a stack of GATES layers. In the forward process of each GATES layer, one step of information propagation is taken place at every node. The detailed formulas and implementations of one GATES layer for ``operation on node'' and ``operation on edge'' search spaces are shown as follows, and the notations are summarized in Table.~\ref{tab:notation}.

\subsubsection{Operation On Node (OON) Search Space}

For the OON case, we take the NAS-Bench-101 search space as an example. In the cell architecture, there is $n_i=1$ input node, and at most $V=7$ nodes. For batch computation, we pad zero columns and rows into the adjacent matrix to ensure that all adjacent matrices are of size $7\times7$, and also add none operations into the corresponding positions in the operation list. The calculation of the $k$-th GATES layer could be written as
\begin{equation}
\begin{aligned}
X^{(0)} &= \mbox{CONCAT}(\tilde{E}, {\bf 0}_{b \times V-n_i \times h_i^{(0)}}, \mbox{dim=1})\\
X^{(k)} &= \sigma(\mbox{EMB}(o) W^{(k)}_o) \odot (A X^{(k-1)} W^{(k)}_x)
\end{aligned}
\label{eq:gates_nasbench}
\end{equation}
where $\tilde{E} = \mbox{repeat}(E, [b, 1, 1]) \in \mathbb{R}^{b \times n_i \times h^{(0)}_i}$, and $E, \mbox{EMB}, W^{(k)}_o, W^{(k)}_x$ are trainable parameters.

In practice, we found that for the OON search space, adding a self-loop of the information propagation would lead to slightly better performance. 
\begin{equation}
    \begin{aligned}
        X^{(k)} &= \sigma(\mbox{EMB}(o) W^{(k)}_o) \odot (\tilde{A} X^{(k-1)} W^{(k)}_x) \\
        \tilde{A} &= A + I
    \end{aligned}
\end{equation}

\subsubsection{Operation On Edge (OOE) Search Space}

For the OOE search spaces, the calculation of a GATES layer could be written as
\begin{equation}
\begin{aligned}
X^{(0)} &= \mbox{CONCAT}(\tilde{E}, {\bf 0}_{b \times V-n_i \times h_i^{(0)}}, \mbox{dim=1})\\
S &= \expand(X^{(k-1)} W^{(k)}_x, 1)\\
X^{(k)} &= \mbox{SUM}(\sum_{d=1}^{n_d} \expand(A, 3) \odot \sigma(\mbox{EMB}(o_d) W^{(k)}_o) \odot S, \mbox{dim=2})
\end{aligned}
\label{eq:gates_enas}
\end{equation}
where $\tilde{E} = \mbox{repeat}(E, [b, 1, 1]) \in \mathbb{R}^{b \times n_i \times h^{(0)}_i}$, and $\expand(A, \mbox{dim})$ denotes the operation to insert a new dimension as dimension \textit{$\mbox{dim}$}.

For the search spaces where there is at most one edge between each pair of nodes (e.g., NAS-Bench-201), the above calculation could be simplified to
\begin{equation}
\begin{aligned}
X^{(0)} &= \mbox{CONCAT}(\tilde{E}, {\bf 0}_{b \times V-n_i \times h_i^{(0)}}, \mbox{dim=1})\\
S &= \expand(X^{(k-1)} W^{(k)}_x, 1)\\
X^{(k)} &= \mbox{SUM}(\expand(A, 3) \odot \sigma(\mbox{EMB}(o) W^{(k)}_o) \odot S, \mbox{dim=2})
\end{aligned}
\label{eq:gates_enas}
\end{equation}

\section{Discussion on Isomorphism}

\subsubsection{GATES maps ismorphic architectures to the same representation}
The encoding process of GATES mimics the actual computation flow: GATES uses multiplicative transforms to mimic the forward process of operations (e.g., \texttt{Conv3x3}), and uses commutative aggregation to mimic actual commutative aggregation of the feature maps.
Naturally, GATES would encode two architectures that give out the same feature map results into the same representation. That is to say, the embedding space of GATES is more meaningful. However, GATES might fail to map non-isomorphic architectures to different representations. And we leave it to future work to ameliorate this problem to further increase the discriminative power of GATES.

In the search spaces which we have experimented with (i.e., NAS-Bench-101, NAS-Bench-201, and ENAS), the combination of feature maps at internal nodes is done via addition operation, which is commutative. Therefore, for encoding the architecture, GATES also uses commutative addition to combine the ``virtual information''.
Note that if the feature map aggregation at some internal node is not commutative (e.g., concatenation), we should use a non-commutative aggregation of the virtual information too.

Another thing to note is that, in the NAS-Bench-101 and ENAS search spaces, the tensors going to the final output node in the cell are concatenated instead of being added together.
Since the concatenation operation is not commutative, different concatenation orders result in different architectures. Nevertheless, in these two search spaces, these models are equivalent through the rearrangement of channels in the following operations.
Therefore, we use addition to aggregate the information at the output node, too. We emphasize that this is a search space specific discussion.

\begin{table}[tb]
\begin{center}
\caption{The Kendall's Tau $\tau$ on 1) NAS-Bench-101 test set 2) the 7-vertex subset of the test set 3) all the isomorphic counterparts of the 7-vertex subset (without de-duplication). The last column shows the sum of the variances of the predicted scores in every isomorphic architectures group, and there are negligible numerical errors in the variance results of GATES and GCN. All the predictors are trained using the hinge pairwise ranking loss on 0.1\% of the training data.}
\label{table:iso-nb101}
\begin{tabular}{c@{\hskip 0.02\linewidth}cccc}
\toprule
\multirow{3}{*}{Encoders} & test set & 7-vertex test set & \multicolumn{2}{c}{7-vertex test set w.o. de-dup.} \\ 
& (42362) & (36064) & \multicolumn{2}{c}{(116102)} \\\cmidrule(lr){2-2}\cmidrule(lr){3-3}\cmidrule(lr){4-5}
& $\tau$   &$\tau$ & $\tau$  & Total Var.  \\\midrule
MLP~\cite{wang2018alphax}  & 0.5272 & 0.5143 & 0.4729 & 43.58    \\
LSTM~\cite{wang2018alphax} & 0.5993 & 0.5877 & 0.5656 & 18.80    \\
GCN~\cite{shi2019multi}    & 0.5790 & 0.5876 & 0.6169 & 1.16E-11 \\\hline
GATES                      & {\bf 0.7789} & {\bf 0.7724} & {\bf 0.7758} & 9.24E-12 \\ \bottomrule
\end{tabular}
\end{center}
\end{table}

We conduct a simple experiment to verify GATES's ability to map isomorphic architectures to the same representation on NAS-Bench-101.
After splitting the train and test sets, there are 36064, 6037, 256, 5 testing architectures with 7, 6, 5, 4 vertices, and 323018, 55973, 2185, 79, 6, 1 training architectures with 7 6, 5, 4, 3, 2 vertices respectively.
Since all isomorphic cell architectures are already removed in NAS-Bench-101, we generate the isomorphic architectures for the 36064 unique testing architectures with 7 vertices, and get 116102 architectures. Among the 36064 architectures, there are 20994 architectures that have isomorphic counterparts. We test different predictors trained with 0.1\% training samples on these 116k architectures and show the results in Table.~\ref{table:iso-nb101}. Since the sequence-based encoding schemes cannot map isomorphic architectures to the same representation, the ranking correlation decreases if no de-duplication procedure is carried out. The last column shows the sum of the variances of the predicted scores in every isomorphic architecture group. We can see that GATES and GCN can map isomorphic architectures to the same representation (a variance of 0 with negligible numeric errors), since only isomorphism-invariant aggregation operations are used in the encoding process.

\begin{figure}[tb]
\begin{center}
\includegraphics[width=0.64\linewidth]{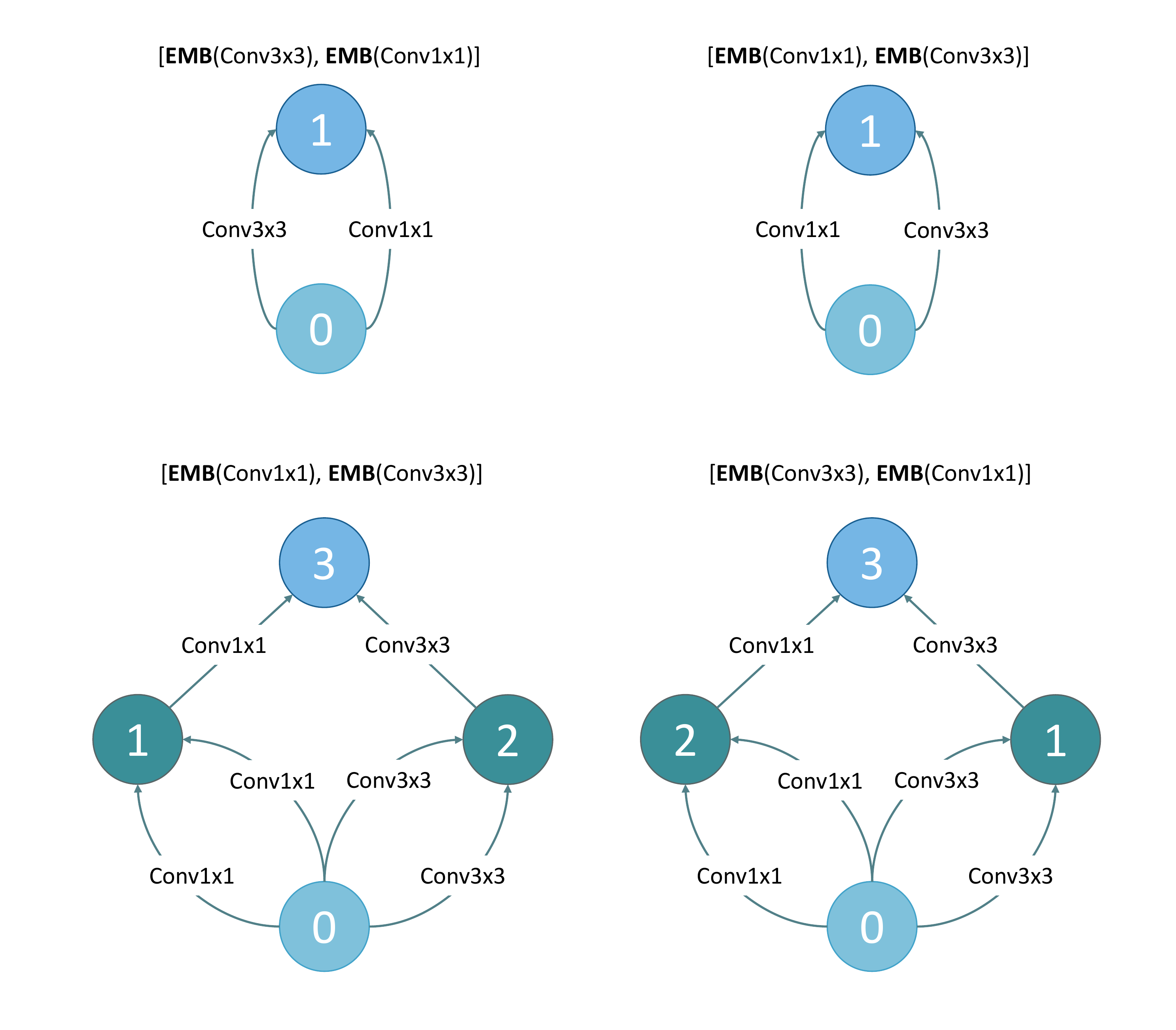}
\caption{An ad-hoc graph-based solution~\cite{guo2019nat} for encoding the architecture in the ENAS search space (an OOE search space) fails to map isomorphic architectures to the same representation. In the upper case, the two architectures are the same graph, but the embeddings of Node 1 differ. This case could be solved by imposing an order of the operations when the two incoming edges come from the same previous node. In the lower case, these two architecture are isomorphic, since the feature map aggregation at Node 3 is a commutative element-wise addition. However, this encoding scheme cannot guarantee to map these two architectures to the same representation, since the original node embeddings already differ at Node 3. The failure to handle the isomorphism is due to the non-commutative characteristics of the concatenation operation.}
\label{fig:guo_gcn_counter_example}
\end{center}
\end{figure}

\subsubsection{Two counter examples of the ad-hoc solution~\cite{guo2019nat}}
Since GCN cannot be directly applied to encoding architectures from the OOE search spaces, a recent study~\cite{guo2019nat} proposes an ad-hoc solution for the ENAS search space. They represent each node by the concatenation of the operation embeddings on the input edges. This solution cannot generalize to search spaces where nodes could have different input degrees. What’s more, since the concatenation operation is not commutative, this encoding scheme could not map isomorphic architectures to the same representation correctly. Fig.~\ref{fig:guo_gcn_counter_example} illustrates two minimal counterexamples.

\section{Setup and Additional Results}

\subsubsection{Setup and Results on NAS-Bench-101}
The setup of all the experiments on NAS-Bench-101 goes as follows. An ADAM optimizer~\cite{kingma2014adam} with learning rate 1e-3 is used to optimize the performance predictors for 200 epochs. And the average of the ranking correlations in the last 5 epochs is reported. The batch size is set to 512. And a hinge pairwise ranking loss with margin 0.1 is used. For the construction of the MLP and LSTM encoder, we follow the serialization method and the model settings in \cite{wang2018alphax}. The MLP is constructed by 4 fully-connected layers with 512, 2048, 2048, and 512 nodes, and the output of dimension 512 is used as the cell's embedding. The embedding and hidden sizes of the LSTM are both set to 100, and the final hidden state is used as the cell's embedding. For the GCN and GATES encoders, we construct the encoder by stacking five 128-dim GCN or GATES layers. 
All the embedding sizes are set to 48, including the operation embedding in GCN, and the operation and information embedding in GATES. For GCN, the average of all the nodes' features is used as the cell's embedding. In GCN with global node~\cite{shi2019multi}, the features of the global node are used as the cell's embedding.

\begin{figure*}[bt]
\includegraphics[width=1.\linewidth]{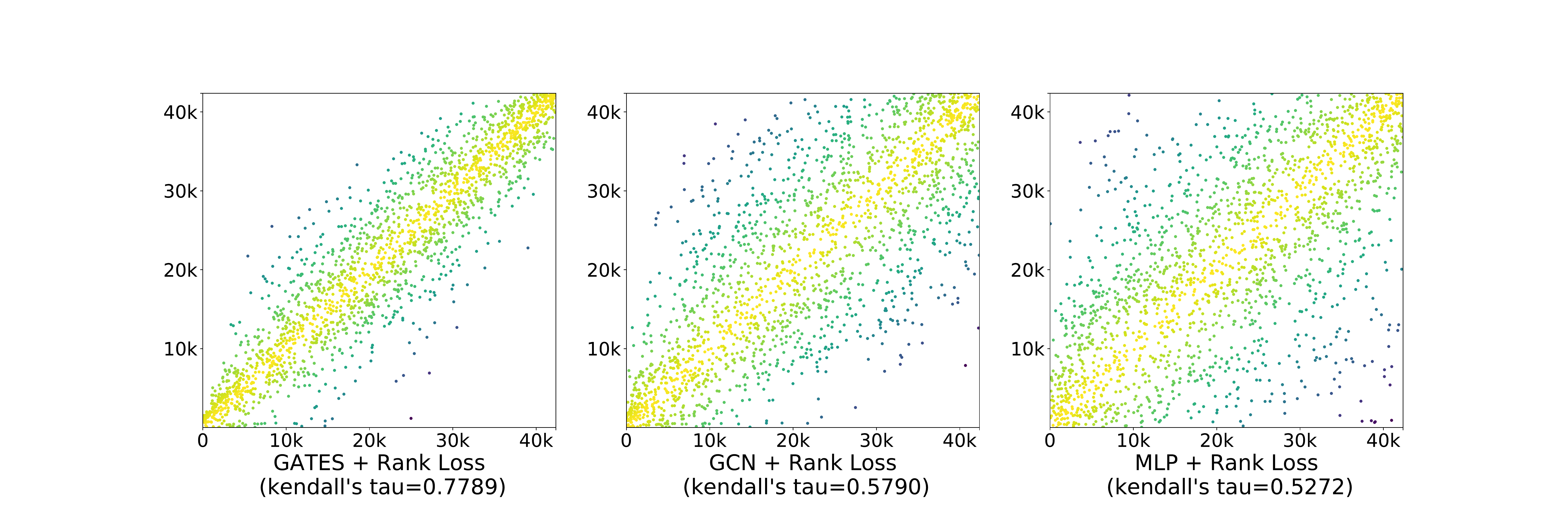}
\caption{NAS-Bench-101: The true rankings (y-axis) and predicted rankings (x-axis) of 2000 architectures among the 42362 testing architectures. 0.1\% training data are used to train these encoders.}
\label{fig:scatter-nb101}
\end{figure*}

Fig.~\ref{fig:scatter-nb101} shows the prediction results on the 42362 testing architectures with different encoders trained on 0.1\% training data.
As can be seen, compared with the GCN and MLP encoders, the predictions of GATES are much more accurate in the sense of ranking correlation.

\subsubsection{Setup and Results on NAS-Bench-201}
The setup of all the experiments on NAS-Bench-201 goes as follows. An ADAM optimizer with learning rate 1e-3 and batch size 512 is used to train the predictors for 200 epochs, and the average of testing Kendall's Taus in the last 5 epochs is reported.

For the sequence-based baselines (MLP and LSTM), we use the 6 elements of the lower triangular portion, excluding the diagonal ones. We use 4 fully-connected layers with 512, 2048, 2048, 512 nodes for the MLP encoder. The embedding size and hidden size of the 1-layer LSTM is set to 100, and the final hidden stage is used as the embedding of the cell architecture. As for GATES, we use a 5-layer GATES encoder without self-loop.

Since GCN encoders could not be directly applied to the OOE search spaces, we implement a line graph solution for applying GCNto encode OOE architectures following these three steps: 1) convert the graph to a line graph; 2) apply an 5-layer GCN; 3) concatenate the node embeddings as the graph representation. The results of this ``Line Graph GCN'' solution are listed in Tab.~\ref{table:gates-nb201}, and are not satisfying enough. We suppose that it is due to that the power of GNN cannot be fully utilized as converting to line graph results in identical adjacent matrices for all NAS-Bench-201 architectures.

\subsubsection{Ablation Study: GATES Layer Number}
\label{sec:exp-ablation-layernum}

\begin{figure*}[tb]
  \begin{center}
  \subfigure[NAS-Bench-101]{
    \includegraphics[width=0.44\linewidth]{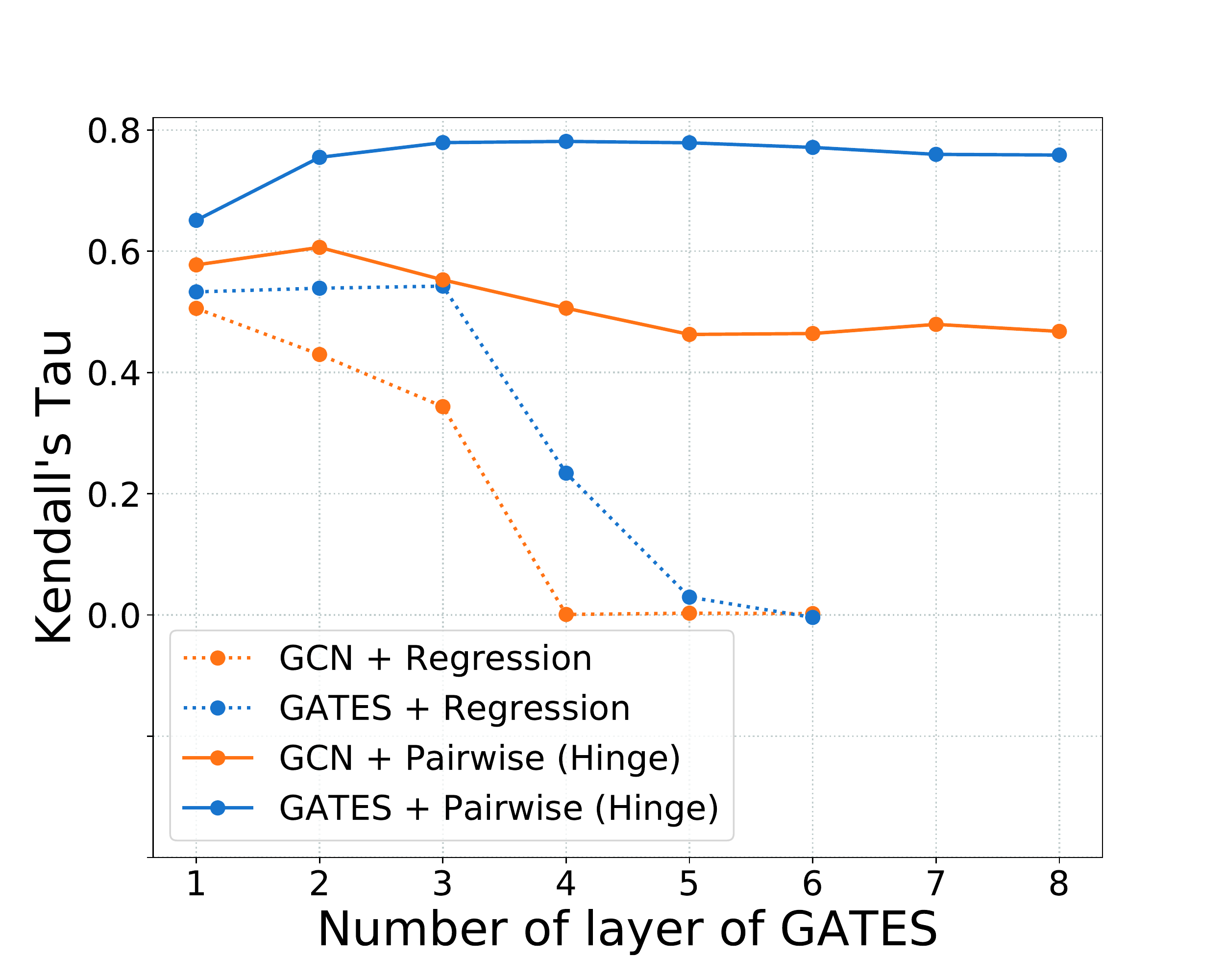} 
    \label{fig:layers_nb101}
  }
  \subfigure[NAS-Bench-201]{
    \includegraphics[width=0.44\linewidth]{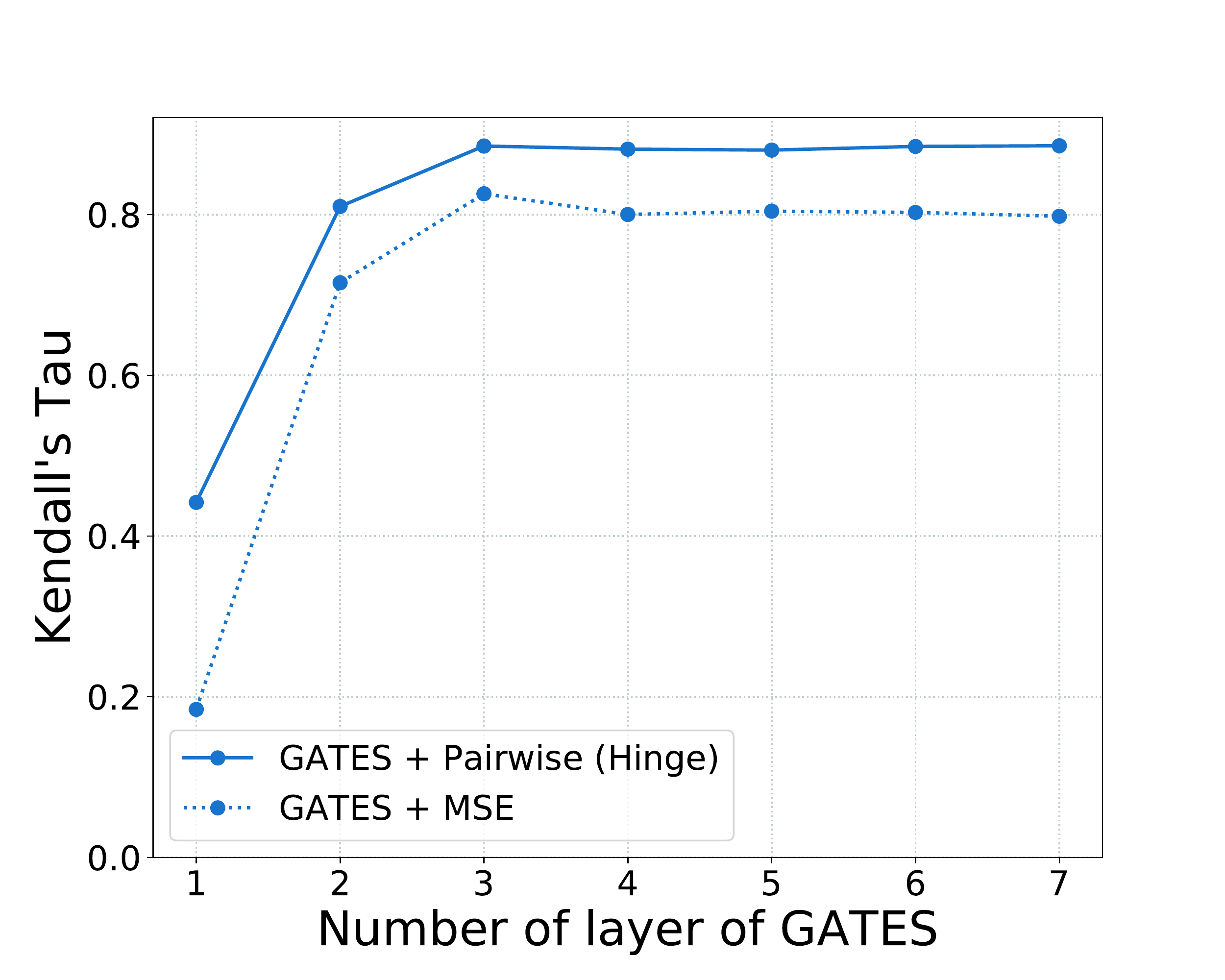} 
    \label{fig:layers_nb201}
  }
  \caption{The  effect of the number of GCN or GATES layers. (a) NAS-Bench-101. The proportion of training samples is 0.1\% (381 training, 42362 testing). (b) NAS-Bench-201. The proportion of training samples is 10\% (781 training, 7812 testing)}
  \label{fig:gates_layers}
\end{center}

\end{figure*}

We show the ablation study of the layer number in the GATES encoders in Fig.~\ref{fig:gates_layers}. We can see that the regression loss fails to instruct the learning of deep GCN and GATES encoders. Even with the ranking loss, the GCN's performance degrades as the layer number increases, while the GATES encoder is more robust.

Another interesting fact is that
a GATES layer number larger than or equal 3 is a good choice on NAS-Bench-201, and as we know, the most common longest path length is 3 too. 
As for NAS-Bench-101, a GATES layer number larger than or equal 4  is a good choice. The longest possible path length on NAS-Bench-101 is 6, but only in a small portion of architectures.
The ablation results match with the ``virtual information flow'' intuition of the GATES design and give evidence of the rationality of using GATES for neural architecture encoding.

\subsubsection{Neural Architecture Search in the ENAS Search Space}
\begin{figure}[tb]
  \begin{center}
    \subfigure[Normal cell]{
      \includegraphics[width=0.75\linewidth]{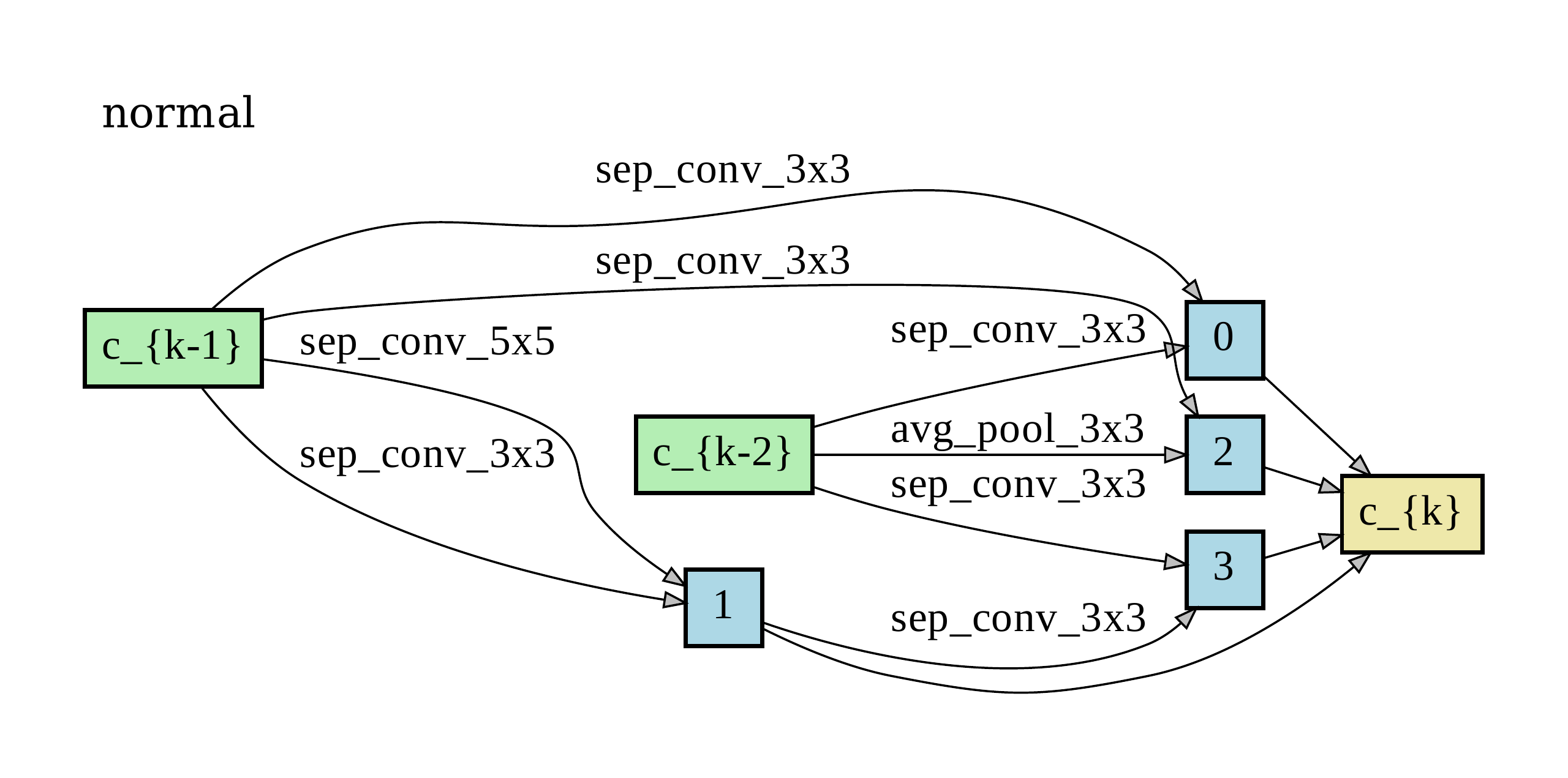} 
    }
    \subfigure[Reduction cell]{
      \includegraphics[width=0.75\linewidth]{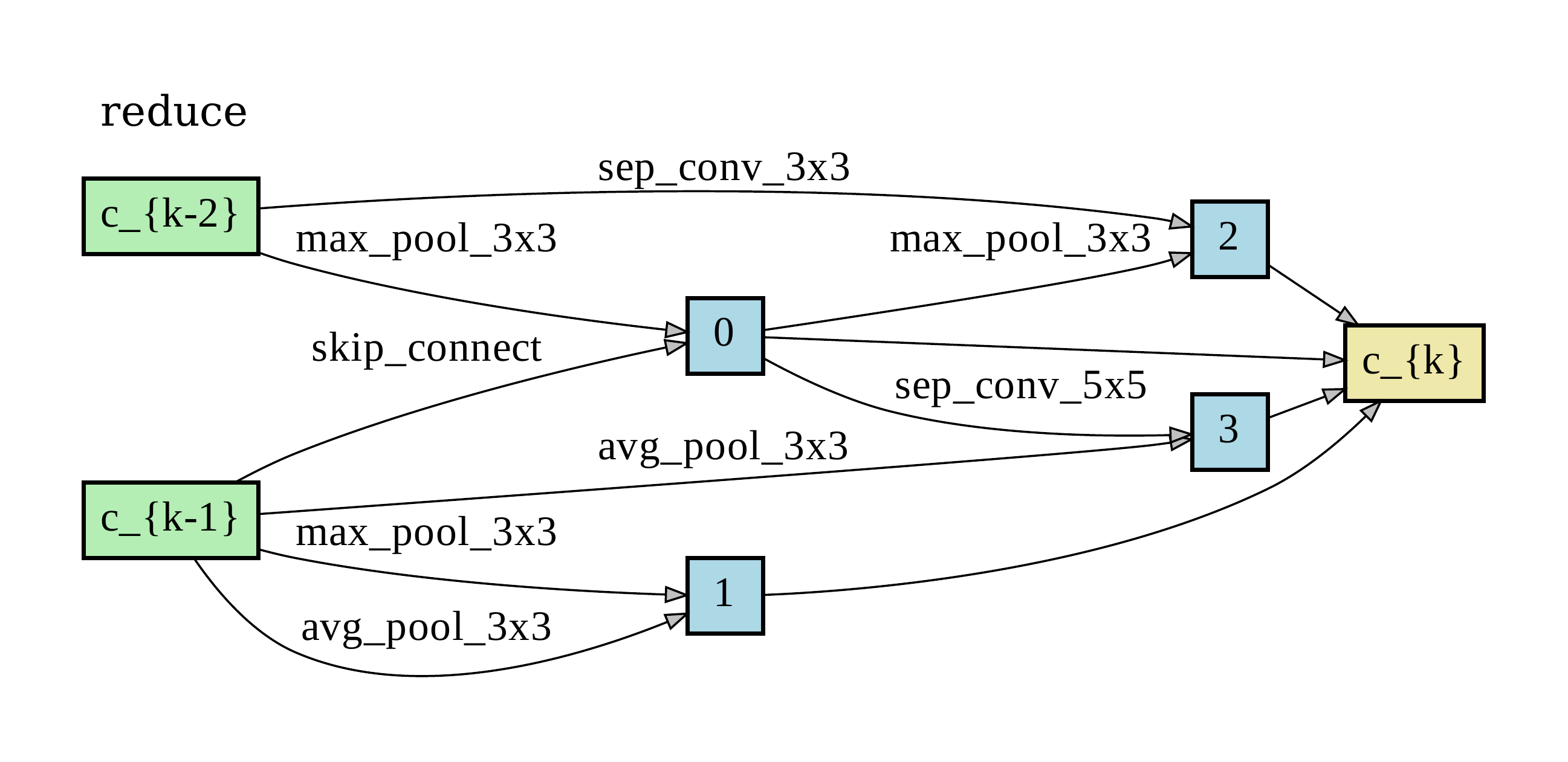} 
    }
    \caption{Discovered cell architectures on CIFAR-10.}
    \label{fig:enas_discover_cell}
  \end{center}
\end{figure}

\begin{table*}[tb]
  \caption{Comparison of NAS-discovered architectures on ImageNet}
  \label{table:imgnet}
  \begin{center}
    \begin{tabular}{ccc}
      \toprule
      Method & Top-1 Test Error (\%) & \#Params (M)\\ \midrule
      NASNet-A~\cite{zoph2016neural}         & 26.0 & 5.3   \\
      AmoebaNet-B~\cite{real2019regularized} & 27.2 & 5.3   \\
      PNAS~\cite{liu2018progressive}                  & 25.8 & 5.1   \\\cmidrule(lr){1-3}
      DARTS~\cite{darts}                       & 26.9 & 4.9   \\
      GHN~\cite{zhang2018graph} & 27.0 & 6.1 \\\hline
      Ours                                   & 24.1 & 5.6 \\ \bottomrule
    \end{tabular}
  \end{center}
\end{table*}

The setup of the predictor training goes as follows. The predictor is constructed by four 64-dim GATES layers. Both the operation and information embedding sizes are set to 32. During the training of the predictor, the total epoch is set to 80, and the batch size is set to 128, and a pairwise hinge loss with margin 0.1 and an ADAM optimizer with learning rate 1e-3 are used.

For the true performance evaluation of the 800 architectures (600 randomly sampled, 200 sampled utilizing the predictor), we train them for 80 epochs using an SGD optimizer with weight decay 3e-4. The learning rate is decayed from 0.05 to 0.001 following a cosine schedule. The base channel number is 16, and the number of layers is 8.

The discovered architecture is shown in Fig.~\ref{fig:enas_discover_cell}. To evaluate the final performance of the discovered cell architecture, we first apply the channel and layer augmentation. Specifically, 20 cells are stacked to construct the network, and the base channel number is increased from 16 to 36. The augmented model is trained for 600 epochs on CIFAR-10 with batch size 128, and the learning rate is decayed from 0.05 to 0.001 following a cosine schedule. The cutout data augmentation with length 16 is used. The weight decay is set to 3e-4, and the dropout rate before the fully-connected classifier is set to 0.1. For other regularization techniques, we follow existing studies~\cite{zoph2018learning,darts} to use auxiliary towers with weight 0.4 and the scheduled drop-path of probability 0.2.

For transferring the discovered architecture to ImageNet, we increase the base channel number to 48 and stack 14 cells to construct the model. The augmented model is trained for 300 epochs with batch size 256, and the learning rate is decayed from 0.1 to 0 following a cosine schedule. The weight decay is set to 3e-5 and auxiliary towers with weight 0.4 is used, no dropout is used. The comparison with a few previous methods is illustrated in Tab.~\ref{table:imgnet}.

\section{Ranking Losses for Predictor Optimization}

The ranking correlation of the performance predictor on unseen architectures is the key to the success of predictor-based NAS. 
Since ranking losses are better surrogates of the ranking measures than the regression loss~\cite{chen2009}, training the performance predictor with ranking losses could lead to better ranking correlation.

We utilize different pairwise and listwise ranking losses for training the predictor~\cite{burges2005learning,shashua2003ranking,xia2008listwise}. The pairwise ranking loss could be written as
\begin{equation}
    \begin{aligned}
    L^p(\tilde{S})= \sum_{i=1}^{N} \sum_{j \in \{j | y_i < y_j\}} \phi(P(a_j), P(a_i))
    \end{aligned}
\end{equation}

We experiment with two different choices of $\phi$. 1) The binary cross entropy function $\phi(s_j, s_i)=\log(1 + e^{(s_j - s_i)})$; 
2) The hinge loss function $\phi(s_j, s_i)= \max(0, m - (s_j - s_i))$, where $m$ is a positive margin.

We also experiment with a pairwise comparator: We construct an MLP that takes the concatenation of two architecture embeddings as input and outputs a score: $s=\mbox{MLP}([E(a_j), E(a_i)]$, and a positive $s$ indicates that $a_j$ is better than $a_i$.  Note that the total-orderness of the architectures is not guaranteed using this comparator. So, we add a simple anti-symmetry regularization term in the training of the comparator. The loss for training the comparator is:
\begin{equation}
    \begin{aligned}
    L^p(\tilde{S})= &\sum_{i=1}^{N} \sum_{j \in \{j | y_i < y_j\}} \max(0, m - \mbox{MLP}([E(a_j), E(a_i)]) \\
    &+ \max(0, m + \mbox{MLP}([E(a_i), E(a_j)]))
    \end{aligned}
\end{equation}

We design the listwise ranking loss following ListMLE~\cite{xia2008listwise}:
\begin{equation}
    L^l(\tilde{S}) = \sum_{U \subset \tilde{S}} \sum_{i=1}^{|U|} \{-P(a^{(i), U}) + \log\sum_{j=i}^{|U|} \exp(P(a^{(j), U}))\}
\end{equation}
where $U$ are subsets of $\tilde{S}$, $|U|$ denotes the size of $U$, $a^{(i), U}$ denotes the architecture whose true performance $y^{(i), U}$ is the $i$-th best in the subset $U$.

\begin{table*}[bt]
  \caption{The Kendall’s Tau of using different loss functions on NAS-Bench-101. The first 90\% (381262) architectures in the dataset are used as the training data, and the other 42362 architectures are used as the testing data. All experiments except ``Regression (MSE) + GCN'' are carried out with GATES encoder.}
  \label{table:rank-nb101}
  \begin{center}
    \resizebox{1.0\textwidth}{!}{
      \begin{tabular}{ccccccccc}
        \toprule
        \multirow{2}{*}{Loss} & \multicolumn{8}{c}{Proportions of 381262 training samples}\\ 
        \cmidrule(lr){2-9} & 0.05\% & 0.1\% & 0.5\% & 1\% & 5\% & 10\% & 50\%  & 100\% \\ \midrule
        Regression (MSE) + GCN$^\dagger$ &0.4536 & 0.5058 & 0.5587 & 0.5699 & 0.5846 & 0.5871 & 0.5901 & 0.5941\\
        Regression (MSE) + GATES$^\dagger$ & 0.4935 & 0.5425 & 0.5739 & 0.6323 & 0.7439 &  0.7849 &  0.8247 & 0.8352\\\hline
        Pairwise (BCE) & 0.7460 & 0.7696 & 0.8352 & 0.8550 & 0.8828 & 0.8913 &  {\bf 0.9006} & {\bf 0.9042} \\
        Pairwise (Comparator) & 0.7250 & 0.7622 &  0.8367 & 0.8540 & 0.8793 & 0.8891 &  0.8987 & 0.9011\\
        Pairwise (Hinge) & {\bf 0.7634} & {\bf 0.7789} & {\bf 0.8434} & {\bf 0.8594} & 0.8841 & {\bf 0.8922} &  0.9001& 0.9030 \\
        Listwise (ListMLE) & 0.7359 & 0.7604 & 0.8312 & 0.8558 & {\bf 0.8852} & 0.8897 &  0.9003 & 0.9009\\\bottomrule
      \end{tabular}
    }
    \begin{minipage}{1.0\textwidth}
        $\dagger$: For the baseline evaluation of regression loss, we use a GCN encoder with 1 layer, and a GATES encoder with 3 layers rather than 5 layers, since training deep GCN or GATES encoder with MSE regression loss is unstable, and often fails to learn anything meaningful. With MSE loss, 1 layer of GCN and 3 layers of GATES achieve the best results among layer number configurations using 0.1\% training data.
    \end{minipage}
  \end{center}
\end{table*}

\subsection{Evaluation of Ranking Losses}
\subsubsection{Setup}

In the experiments of evaluating the ranking losses, the training settings and the construction of the GATES model are the same as in the evaluation of GATES. 
One exception is that, for the listwise ranking loss (ListMLE), we train the predictor for 80 epochs (list length is 4), since the training converges much faster with the listwise ranking loss. Still, the average of the ranking correlations in the last 5 epochs is reported.

The evaluation of the comparator-based ranking loss is a little different than other ranking losses. For other ranking losses, we can calculate the ranking correlation between the predicted scores $\mbox{P}(a)$ and the true accuracies. However, a comparator trained using the comparator-based ranking loss must take a pair of architectures as the input and output a comparison results. Therefore, for evaluating the performance of the comparator,
we run the randomized quick-sort procedure with the comparator to get the predicted rankings of the testing architectures. Since the comparator might not be a proper total order operator, different choices of the random pivots in randomized quick-sort could lead to different sorted sequences. Therefore, we run randomized quick-sort with 3 different random seeds, and report the average Kendall's Tau. In practice, we find that the Kendall's Taus calculated using different random seeds are very close. For example, three tests with random seed 1, 12 and 123 of the predictor trained on the whole training set give the Kendall's Taus of 0.90106, 0.90107 and 0.90113, respectively.

\addtolength{\tabcolsep}{1pt}
\begin{table*}[tb]
\caption{The Kendall’s Tau of using different encoders and loss functions on NAS-Bench-201. The first 50\% (7813) architectures in the dataset are used as the training data, and the other 7812 architectures are used as the testing data}
\label{table:gates-nb201}
\begin{center}

\begin{tabular}{cccccc}
\toprule
\multirow{2}{*}{Encoder} & \multicolumn{5}{c}{Proportions of 7813 training samples}\\ 
\cmidrule(lr){2-6} & 1\% & 5\% & 10\% & 50\% & 100\% \\\midrule
MLP + Regression (MSE)$^\dagger$ & 0.0646 & 0.1520 & 0.2316 & 0.5156 & 0.6089  \\
  LSTM + Regression (MSE) & 0.4405 & 0.5435 & 0.6002 & 0.8169 & 0.8614\\
  Line Graph GCN + Regression (Hinge) &  -0.0481 &  0.3376 & 0.4988 & 0.6609 & 0.7006\\
GATES + Regression (MSE) & {\bf 0.6823} & {\bf 0.7528} & {\bf 0.8042} & {\bf 0.8950} & {\bf 0.9115}\\\hline
MLP + Pairwise (Hinge)  &  0.0974 & 0.3959 & 0.5388 & 0.8229 & 0.8703\\
  LSTM + Pairwise (Hinge) & 0.5550 & 0.6407 & 0.7268 & 0.8791 & 0.9002\\
  Line Graph GCN + Pairwise (Hinge) &   0.5063 & 0.6822 & 0.7567 & 0.8676 & 0.9002 \\ 
  GATES + Pairwise (Hinge) & {\bf 0.7401} & {\bf 0.8628} & {\bf 0.8802} & {\bf 0.9192} & {\bf 0.9259}\\\bottomrule

\end{tabular}
\begin{minipage}{1.0\textwidth}
$\dagger$: For the baseline evaluation of MSE regression loss with MLP and Line Graph GCN encoders, we use a learning rate of 1e-4, since we find out that these encoders cannot be learned with a learning rate of 1e-3.
\end{minipage}
\end{center}
\end{table*}
\addtolength{\tabcolsep}{-1pt}

\subsubsection{Results on NAS-Bench-101}
We train GATES-powered predictors with four types of ranking losses: 1) Pairwise loss with binary cross-entropy $\phi$. 2) Pairwise loss with a hinge loss function $\phi$. 3) Pairwise comparator loss. 4) Listwise (ListMLE).
Table~\ref{table:rank-nb101} shows the comparison of using different losses to train the predictors on NAS-Bench-101. Compared with the regression loss, ranking losses bring consistent improvements.
The performances of different ranking losses are close, and the pairwise hinge loss is a good choice. We also find that training with regression loss requires a smaller learning rate and longer time to converge, and does not work well with 
deep GCN or GATES models.

\subsubsection{Results on NAS-Bench-201}
Table~\ref{table:gates-nb201} shows the comparison of using regression and ranking losses to train the predictors on NAS-Bench-201. We can see that training using ranking losses leads to better-correlated predictors consistently.

\end{document}